\documentclass[12pt,twoside]{elsarticle}

\usepackage{array}
\usepackage{tabularx}
\usepackage{hyperref}
\usepackage{amsfonts} 
\usepackage{breqn}
\usepackage{multirow}
\usepackage{pdflscape}
\usepackage{float}
\usepackage{caption}
\usepackage{xcolor}
\usepackage[ruled,vlined]{algorithm2e}
\usepackage{amssymb}
\usepackage{fancyhdr}
\usepackage{etoolbox}
\usepackage[a4paper, left=2.5cm, right=2.5cm, top=3.5cm, bottom=2.5cm]{geometry}

\newcommand{\EvenLeft}{}
\newcommand{\EvenRight}{}
\newcommand{\OddLeft}{}
\newcommand{\OddRight}{}

\hypersetup{
 colorlinks,
 citecolor=black,
 filecolor=black,
 linkcolor=black,
 urlcolor=black
}
\definecolor{darkred}{rgb}{0.9,0.1,0.1}
\biboptions{sort&compress}
\hypersetup{
 colorlinks,
 citecolor=black,
 filecolor=black,
 linkcolor=black,
 urlcolor=black
}

\begin{document}

\begin{frontmatter}

\title{\textbf{Residual-Based Attention in Physics-Informed Neural Networks\tnoteref{pubnote}}}
\author[inst1,label1]{Sokratis J. Anagnostopoulos}
\author[inst2, label1]{Juan Diego Toscano}
\author[inst1]{\\ Nikolaos Stergiopulos}
\author[inst2,inst3]{George Em Karniadakis}

\affiliation[inst1]{organization={Laboratory of Hemodynamics and Cardiovascular Technology, EPFL},
   city={Lausanne},
   postcode={1015}, 
   state={VD},
   country={Switzerland}}
   
\affiliation[inst2]{organization={Division of Applied Mathematics, Brown University},
   city={Providence},
   postcode={02912}, 
   state={RI},
   country={USA}}
   
\affiliation[inst3]{organization={School of Engineering, Brown University},
   city={Providence},
   postcode={02912}, 
   state={RI},
   country={USA}}

\tnotetext[pubnote]{Published in \textit{Computer Methods in Applied Mechanics and Engineering}: \\ \hspace*{1.5em} \href{https://doi.org/10.1016/j.cma.2024.116805}{doi:10.1016/j.cma.2024.116805}.
Updated to match publication.}

\fntext[label1]{The first two authors contributed equally to this work}
\begin{abstract}
Driven by the need for more efficient and seamless integration of physical models and data, physics-informed neural networks (PINNs) have seen a surge of interest in recent years. However, ensuring the reliability of their convergence and accuracy remains a challenge. In this work, we propose an efficient, gradient-less weighting scheme for PINNs that accelerates the convergence of dynamic or static systems. This simple yet effective attention mechanism is a bounded function of the evolving cumulative residuals and aims to make the optimizer aware of problematic regions at no extra computational cost or adversarial learning. We illustrate that this general method consistently achieves one order of magnitude faster convergence than vanilla PINNs and a minimum relative $L^{2}$ error of $\mathcal{O}(10^{-5})$, on typical benchmarks of the literature. The method is further tested on the inverse solution of the Navier-Stokes within the brain perivascular spaces, where it considerably improves the prediction accuracy. Furthermore, an ablation study is performed for each case to identify the contribution of the components that enhance the vanilla PINN formulation. Evident from the convergence trajectories is the ability of the optimizer to effectively escape from poor local minima or saddle points while focusing on the challenging domain regions, which consistently have a high residual score. We believe that alongside exact boundary conditions and other model reparameterizations, this type of attention mask could be an essential element for fast training of both PINNs and neural operators.
\end{abstract}


\end{frontmatter}

\newpage

\section{Introduction}
\label{p2:Introduction}

Physics-informed neural networks (PINNs) have emerged as an alternative to traditional numerical methods for solving partial differential equations (PDEs) in forward and inverse problems \cite{raissi2019physics}. In the PINN approach, a neural network is trained to approximate the solution of a PDE by minimizing a composite loss function that includes terms related to the physical laws and data from initial boundary conditions, simulations, or experiments \cite{cai2021physics}. Keeping a balance between the various and diverse loss terms is crucial to avoid convergence and accuracy issues, especially when dealing with highly nonlinear, multi-scale, or chaotic behavior problems \cite{wang2022respecting,mcclenny2020self,es2023optimal}. To address this imbalance, researchers have developed several approaches that can be roughly classified into three categories: neural network modifications, expression transformations, and adaptive weighting strategies.

A neural network modification addresses the multi-layer perceptron optimization capabilities via domain decompositions \cite{kopanivcakova2023enhancing}, model reparametrizations \cite{wang2021understanding, salimans2016weight}, input dimension expansions \cite{guan2022dimension,wang2021eigenvector}, sequential learning \cite{krishnapriyan2021characterizing,wight2020solving}, or adaptive activation functions \cite{jagtap2020adaptive}. On the other hand, expression transformations attempt to simplify the problem by converting the physical laws into their equivalent forms or re-formulating soft constraints into strict ones. For instance, by using alternate or auxiliary physical laws, it is possible to reduce the number of equations or decrease the PDE order \cite{jin2021nsfnets,basir2022investigating}, which eases the optimization process. Similarly, by exactly imposing the boundary conditions, the number of losses can be reduced, and the optimizer can focus on a single condition \cite{sukumar2022exact,leake2020deep,lu2021physics}. 

Adaptive weighting strategies address the loss imbalances by iteratively altering the contribution of particular terms or regions of the domain during the training process. This modification can be done indirectly by adaptively resampling points in crucial areas \cite{lu2021deepxde,wu2023comprehensive,zapf2022investigating} or directly by assigning multipliers that adjust the contribution of each error function. For instance, Wang et al. \cite{wang2021understanding} proposed a learning rate annealing algorithm that balances the losses' weights based on their back-propagated gradients. Wang et al. \cite{wang2022respecting} also defined a causal parameter for time-dependent problems that force the model to learn sequentially based on the time steps. Global multipliers generally modify the average contribution of each loss term \cite{wang2021understanding,wang2022respecting,xiang2022self,liu2021dual}. 
Similarly, these multipliers can be applied locally (i.e., per training point); McCLenny et al. \cite{mcclenny2020self} proposed a self-adaptive (SA) method where the individual loss weights are trained via adversarial training. Zhang et al. \cite{zhang2023dasa} extended this concept and proposed a differentiable adversarial self-adaptive (DASA) pointwise weighting scheme that uses a subnetwork to find the optimal multipliers. Basir et al. presented a physics and quality-constrained artificial neural network (PECANN) that computes the local weights based on the residuals of the constraints (i.e., initial and boundary conditions) and the augmented Lagrangian method \cite{basir2022physics}. PECANN was extended in subsequent studies using adaptive (PECANN-AL) and global Lagrange multipliers \cite{basir2022investigating,basir2023adaptive}. Likewise, Son et al. \cite{son2023enhanced}  proposed an augmented lagrangian relaxation method for PINNs (AL-PINNs), using the initial and boundary conditions as constraints for the optimization problem of the PDE residual. These weighting strategies enhance the capabilities of PINNs and enable them to address challenging problems in diverse fields. Nevertheless, their implementation can be expensive since they may require coupling auxiliary networks \cite{zhang2023dasa}, training additional multipliers \cite{mcclenny2020self}, or computing and processing gradients \cite{wang2021understanding}. Moreover, some of these approaches grow unboundedly \cite{mcclenny2020self, basir2022investigating}, which may result in over-refining some regions while ignoring others \cite{zhang2023dasa}.

To address these problems, we propose a residual-based attention (RBA) scheme that efficiently computes adaptive weights for the training points using the residuals of their given losses. The loss residuals contain information about the high error regions, so the obtained weights work as an attention mask and help the optimizer focus on capturing the spatial or temporal characteristics of the given problem. The suggested weighting approach considers the neural network training dynamics and adapts itself during the training process.

This paper is organized as follows. In Section~\ref{p2:Methods}, we provide an overview of PINNs, introduce the RBA weighting scheme, and review additional modifications that enhance the model performance. Then, in Section~\ref{p2:Results}, we present the implementation of our method through two benchmark problems. To further evaluate our model performance in real-world applications, we combine RBA with artificial intelligence velocimetry (AIV) \cite{boster2023artificial,zhang2023artificial} to infer high-resolution velocity and pressure fields in the cerebrospinal fluid
in the perivascular brain space using real data. We conclude each problem with an ablation study that helps us interpret the relative contribution of each modification. Finally, in Section~\ref{p2:Summary}, we summarize the findings and discuss our future work. In the appendix, we include details regarding the implementation. The relevant code is available on Github: https://github.com/soanagno/rba-pinns.

\section{Methods}
\label{p2:Methods}

\subsection{Physics-Informed Neural networks}

In the context of Physics-Informed Neural Networks (PINNs), a general partial differential equation (PDE) can be expressed as:

\begin{equation}
\mathcal{D}\{u(x,t)\} = f(x,t),
\end{equation}

\noindent where \(u(x,t)\) is the unknown function we wish to approximate, \(\mathcal{D}\) denotes the differential operator, and \(f(x,t)\) is a given source or forcing function that introduces external influences to the system. The differential operator \(\mathcal{D}\) depends on the specific PDE under consideration. For instance, in the case of the heat equation, \(\mathcal{D} = \frac{\partial}{\partial t} - \kappa \nabla^2\), where \(\kappa\) is the thermal diffusivity and \(\nabla^2\) is the Laplace operator.

Additionally, the problem may be constrained by several types of boundary conditions. These conditions are typically categorized into Dirichlet, Neumann, Robin, or Periodic, and are represented as follows:
\begin{alignat}{3}
&u(x_b,t) = g_d(x_b,t), &\text{(Dirichlet)}\\
&\frac{\partial u}{\partial n}(x_b,t) = g_n(x_b,t), &\text{(Neumann)}\\
&\alpha u(x_b,t) + \beta \frac{\partial u}{\partial n}(x_b,t) = g_r(x_b,t), &\text{(Robin)}\\
&u(x_b+L,t) = u(x_b,t), &\text{(Periodic)}
\end{alignat}

\noindent where \(x_b\) denotes the boundary points. The Dirichlet condition defines the function value at the boundary, the Neumann condition specifies the normal derivative at the boundary, and the Robin condition is a weighted combination of the function and its normal derivative. For the Periodic condition, the function value repeats after a certain length \(L\). 

The loss function in PINNs is designed to encompass the deviation of the neural network solution from the initial/boundary conditions, data and the PDE itself. The influence of these components in the loss function is balanced using weighting multipliers \(w_{ic}\), \(w_{bc}\), {\(w_{d}\)} and \(w_r\):
\begin{equation}
\mathcal{L} = w_{ic} \mathcal{L}_{ic} + w_{bc} \mathcal{L}_{bc} + {w_d \mathcal{L}_{d}} + w_r\mathcal{L}_{r},
\label{p2:eq:loss}
\end{equation}

\noindent where \(\mathcal{L}_{ic}\), \(\mathcal{L}_{bc}\), and \(\mathcal{L}_{r}\) represent the residuals associated with the initial conditions, boundary conditions, {any available data} and the PDE, respectively. These terms are given by:
\begin{equation}
\mathcal{L}_{ic} = \frac{1}{N_{\text{ic}}} \sum_{i=1}^{N_{\text{ic}}} ( u_{NN}(x_{i,\text{ic}}, t_{i,\text{ic}}) - u_{i,\text{ic}} )^2
\label{p2:eq:ic}
\end{equation}

\begin{equation}
\mathcal{L}_{bc} = \frac{1}{N_{\text{bc}}} \sum_{i=1}^{N_{\text{bc}}} ( \mathcal{B}\{u_{NN}(x_{i,\text{bc}}, t_{i,\text{bc}}), u_{i,\text{bc}}\} )^2
\end{equation}

\begin{equation}
\mathcal{L}_{r} = \frac{1}{N_{\text{r}}} \sum_{i=1}^{N_{\text{r}}} ( \mathcal{R}\{u_{NN}(x_{i,\text{r}}, t_{i,\text{r}}), x_{i,\text{r}}, t_{i,\text{r}}\} )^2,
\end{equation}

\noindent where $u_{NN}$ is the neural network prediction, {$u_{i,j}$} is the expected value at collocation point $i$ with spatio-temporal coordinates $(x, t)$, $\mathcal{B}$ is the boundary condition operator, $\mathcal{R}$ is the PDE residual operator and each $N$ is the number of collocation points of each set. {Note that Eq. \ref{p2:eq:ic} can also apply to data.}

In general, {a loss term $\mathcal{L}_{j}$} takes the form:

\begin{equation}
{\mathcal{L}_{j}} = \langle (\lambda_{i,j} \cdot e_{i,j}(u_{NN}))^{n} \rangle_{i}, \ \ for \ \ j \in \{ic, bc, {d}, r\} 
\label{p2:eq:loss_term}
\end{equation}

\noindent where $\langle \cdot \rangle$ is the mean operator, $\lambda_{i,j}$, is an attention mask of multipliers \cite{mcclenny2020self} and $e_{i, j}$ is a function of the neural network $u_{NN}$, associated with the error or residual between the prediction and the ground truth.

Each of these terms encourages the neural network to conform to the respective physical laws or initial/boundary conditions encoded in the problem. Note that the weighting multipliers can be global (e.g., scalars $w$ balancing each individual loss term) or local (e.g., vectors $\lambda$ balancing each collocation or boundary point). In the most basic PINN definition, $w=\lambda=1$ for all loss terms.

The training of the neural network weights is performed using {a version} of gradient descent (GD). In each iteration \(k\), an update is applied to a network weight \(\theta\) as follows:

\begin{equation}
\theta^{k+1} = \theta^k - \eta \cdot \nabla_{\theta} \mathcal{L},
\end{equation}

\noindent where \(\eta\) is the learning rate, and \(\nabla_{\theta} \mathcal{L}\) is the gradient of the loss function with respect to \(\theta\). {Each iteration updates the parameters $\theta$ iteratively}, pushing the solution towards satisfying the given PDE and boundary/initial conditions. {For batch GD-methods, each epoch consists of multiple mini-batch iterations.}

\subsection{Residual-based attention (RBA) scheme}

  One of the inherent challenges in obtaining accurate results with PINNs is that the residuals at key collocation points can get overlooked by the mean calculation of the objective function (Eq. \ref{p2:eq:loss_term}). Consequently, despite a decrease in total loss during training, certain spatial or temporal characteristics might not be fully captured. This issue becomes particularly pronounced in multiscale problems, where such troublesome areas could not only result in a lack of detail but could also impair the flow of important information from the initial and boundary conditions into the domain of interest, thereby hindering convergence. Although mini-batching techniques present a potential solution to this issue, the difficult task of a priori identifying the problematic regions remains.

  A workaround for this problem is selecting a set of weighting multipliers, either global or local, which can be adjustable during training. The primary 
  aim of the global multipliers is to balance different terms of the loss function, while local multipliers aim at weighting the influence of specific collocation points of the domain. Establishing a systematic rule for updating these multipliers can significantly aid the optimizer, with several successful strategies documented in the literature, demonstrating major improvements to the vanilla PINNs performance \cite{wang2022respecting,mcclenny2020self,zhang2023dasa,son2023enhanced}.

  Furthermore, as opposed to classical numerical methods, which can guarantee stability, the training convergence of PINNs is affected by the evolution of residuals, which will have some degree of stochasticity, leading to a corresponding degree of instability. Moreover, convergence techniques applied on non-convex loss landscapes can only guarantee a solution as a local optimum, while there might exist other better local (or global) optima.  Driven by the aforementioned challenges, the initial objective of this work is the development of a simple, gradient-less attention mask based on the rolling history of the cumulative residuals to enhance the focus of the optimizer. The proposed scheme aims to increase attention to the challenging regions of information propagation in both space and time dimensions defining the problem.
  
  Based on the exponentially weighted moving average (EWMA) formulation of classic optimizers like RMSprop and Adam \cite{kingma2014adam}, the update rule for the proposed residual-based multipliers for any collocation point $i$ on iteration $k$ is given by:

  \begin{equation}
  \lambda_i^{k+1} \leftarrow \gamma\lambda_i^{k}+\eta^* {\frac{\left| e_i \right|}{\;\,\,\lVert e \rVert_{\infty}}}, \ \ i \in \{0, 1, ..., N\},
  \label{p2:Update_RBA}
  \end{equation}
      
    \noindent where $N$ is the number of collocation points, $e_i$ is the residual of each loss term for point $i$, $\gamma$ is the decay parameter and $\eta^*$ is the learning rate, respectively. Note that the learning rate of the optimizer $\eta$ and the learning rate of the weighting scheme $\eta^*$ are two different hyperparameters.
  
  This is a convergent linear homogeneous recurrence relation. Given that $\frac{\left| e_i \right|}{\;\;\;\lVert e \rVert_{\infty}} \in [0, 1]$ and $\lambda_i^{1} \neq 0, \forall i \in \{0, 1, ..., N\}$, its bounds are given by:

  \begin{equation}
  \lambda_i^{k} \in (0, \frac{\eta^*}{1-\gamma}],\ \ as\ k \rightarrow \infty
  \label{p2:eq:bound}
  \end{equation}
  
  These local multipliers scale with the normalized cumulative residuals, but they also behave dynamically during training due to the decay $\gamma$ as follows: since $\gamma$ lies between 0 and 1, $\lambda^{k}$ will gradually decrease in its contribution to $\lambda^{k+1}$, as k increases. Simultaneously, $\eta^* \frac{\left| e_i \right|}{\;\;\;\lVert e \rVert_{\infty}}$ is being added in each iteration, leading to an eventual equilibrium. However, as different points $i$ may be targeted depending on the stage of the training process, the distribution of $\lambda_i$ will dynamically vary across the domain of interest.
  
  The main advantages of the proposed residual weighting scheme are summarized as follows:
  \begin{enumerate}
  \item Deterministic operation scheme, bounded by the $\gamma$ and $\eta^*$ parameters, which ensure the absence of exploding multipliers. 
  \item No training or gradient calculation is involved, leading to negligible additional computational cost.
  \item Scaling with the cumulative residuals guarantees increased attention on the solution fronts where the PDEs are unsatisfied in both spatial and temporal dimensions.
  \end{enumerate}
 
 The modified training process for an indicative Gradient Descent optimization method is outlined in Algorithm~\ref{p2:update_rule}. We define the RBA weights as $\lambda_{i,j}$ and initialize them with 0 values for all collocation points, {so that at $k=1$ we get $\lambda_{i}^1=\eta^*\frac{\left| e_i \right|}{\;\;\;\lVert e \rVert_{\infty}}$. We also test the option where $\eta^*=1$ only for $k=1$, so that $\lambda_i^1 = \frac{\left| e_i \right|}{\;\;\;\lVert e \rVert_{\infty}}$. Both variations yield similar performance}. To avoid having very small $\lambda$ values in problems where outliers or noisy measurements might be present, an offset term $\lambda_{o}$ may be added, which works as a lower bound such that $\lambda_{i,j}\in(\lambda_{o},\frac{\eta^*}{1-\gamma}+\lambda_{o}]$. This was only used in the case of Section \ref{p2:3DNS}.

  \vspace{5mm}
  \begin{algorithm}[H]
  \label{p2:update_rule}
  \SetAlgoLined
  \KwData{PDE/Boundary collocation points $i$, learning \\ rate $\eta$, RBA learning rate $\eta^*_0$, decay $\gamma$, offset term $\lambda_{o}$}
  \KwResult{Optimized network parameters $\theta_{i}, \ i \in \{0, 1, ..., N\}$}
 
 \textbf{Initialization}: Xavier \cite{glorot2010understanding} $\forall$ $\theta_{i}$ and 0 $\forall$ $\lambda_{i, j}$, {$init=1 \ or \ 2$}\;

 \For{training iterations $k$}{

  \For{each loss term $j$}{
    Get the PDE and boundary residuals $e_{i,j}$\;
        {
        \eIf{$init=2$ and $k=1$}{
            $\eta^* = 1$;
        }{
            $\eta^* = \eta^*_0$;
        }
        }
    $\lambda_{i,j}^{k+1} \leftarrow \gamma\lambda_{i,j}^{k}+\eta^*\frac{\left| e_i \right|}{\;\;\;\lVert e \rVert_{\infty}} $\;
    $\lambda_{i,j}^{*} \leftarrow \lambda_{i,j}^{k+1} + \lambda_{o}$
    }
  $\mathcal{L} = \sum_{j=1}^{n} \langle (\lambda^{*}_{i,j} \cdot e_{i,j})^{2} \rangle_{i}$\;
  $\theta^{k+1} = \theta^{k} - \eta \cdot \nabla_{\theta}\mathcal{L}$\;

 }
 
 \caption{Residual-based attention Gradient Descent}
 \end{algorithm}

\subsection{Additional PINN enhancements}
\label{p2:enhancements}

\subsubsection{Modified multi-layer perceptrons}
\label{p2:MMLP}

For the design of the network architecture, we leverage the recently introduced modified multi-layer perceptron (mMLP) \cite{wang2021understanding}, which draws its inspiration from the attention mechanisms prominent in transformers \cite{vaswani2017attention}. The mMLP aims to augment the efficacy of PINNs by embedding the input variables $x$ into the hidden states of the network. Initially, these inputs are encoded in a feature space by employing two distinct encoders, $U$ and $V$, given by:

\begin{equation}
U = \sigma(x^{0}\theta^U + b^U), \quad V = \sigma(x^{0}\theta^V + b^V)
\end{equation}

 The encoders are then assimilated within each hidden layer of a conventional MLP by point-wise multiplication. Thus, each forward pass becomes:

\begin{equation}
\alpha^{l}(x) = \alpha^{l-1}(x)\theta^{l} + b^{l}, \quad for \ \ l \in \{1, 2, ..., layers\}
\end{equation}

\begin{equation}
\alpha^{l}(x) = \sigma(\alpha^{l}(x))
\end{equation}

\begin{equation}
\alpha^{l}(x) = (1-\alpha^{l}(x)) \odot U + \alpha^{l}(x) \odot V,
\end{equation}

\noindent where $x$ is the input, $\alpha^{l}$ and $\theta^{l}$ are the neurons and weights of layer $l$, $\sigma$ is the activation function and $\odot$ is the element-wise product. 

\subsubsection{Weight Normalization} 

Weight Normalization is a reparameterization technique that accelerates convergence in PINNs \cite{raissi2020hidden}. In this scheme, the weight vectors' length and direction are decoupled to be trained separately \cite{salimans2016weight}. The computation of each neuron output is defined as follows,
\begin{align}
    \alpha&=\sigma(\theta \cdot x + b)\\
    \theta&=\frac{g}{\lVert\mathbf{v}\rVert_{2}}\mathbf{v}\text{,}
    \label{p2:eq_wn}
\end{align}

\noindent where $\alpha$ is the neuron output, $\sigma$ is the activation function, $x$ is the input vector, $\theta$ is a weight vector, and $b$ is the bias. As shown in equation~\ref{p2:eq_wn}, the weight vector $\theta$ is redefined in terms of new trainable parameters, $\mathbf{v}$ (direction) and $g$ (length). Notice that $||\theta||=g$, so this reparameterization allows us to decouple the weight's length and direction, which speeds up the model convergence. Since $g$ is a scalar, this modification induces minimal computational overhead \cite{salimans2016weight}.

\subsubsection{Exact imposition of boundary conditions} 
\label{p2:ExactBCs}
The inexact imposition of boundary conditions can harm the performance and training stability of neural networks \cite{dong2021method,wang2021understanding,chen2020comparison}. Therefore, for our experiments, we consider exactly imposing Dirichlet or Periodic boundary conditions as follows:

\textit{Dirichlet Boundary Conditions}. A recent study by Sukumar et al. \cite{sukumar2022exact} showed how to impose Dirichlet, Neumann, and Robin boundary conditions in PINNs by using approximate distance functions (ADF). The constrained expression for Dirichlet boundary conditions is defined in Eq. \ref{p2:ADF}.

\begin{equation}
 u(\mathbf{x})=g(\mathbf{x})+\phi(\mathbf{x})u_{NN}(\mathbf{x}),
 \label{p2:ADF}
\end{equation}

\noindent where $g(\mathbf{x})$ is a function that satisfies $u$ along the boundaries, $u_{NN}(\mathbf{x})$ is the output of our neural network, and $\phi(\mathbf{x})$ is a composite distance function that equals zero when evaluated on the boundaries. If the boundary is composed of $M$ partitions $[S_1,...S_M]$, the composite distance function for Dirichlet boundary conditions can be described as:

 \begin{align}
  \phi(\phi_1,\phi_2,...,\phi_M)=\Pi_i^M \phi_i,
 \end{align}

\noindent where $[\phi_1,..,\phi_M]$ are the individual distance functions. Notice that if $\mathbf{x}\in S_i$, then $ \phi(\mathbf{x})=0 $ and the neural network approximation exactly satisfies the boundary conditions, i.e., $u(\mathbf{x})=g(\mathbf{x})$.
 
\textit{Periodic Boundary Conditions.} Similarly, it is possible to enforce periodic boundary conditions as hard constraints strictly by constructing Fourier feature embeddings of the input data \cite{wang2022respecting,dong2021method}. In particular, the periodic constraint of a smooth function $u(x)$ can be encoded in a neural network using a one-dimensional Fourier feature embedding (See equation~\ref{p2:fourier1d}).
\begin{equation}
 v(x,y)=[1,cos(\omega_x x),sin(\omega_x x),...,cos(m\omega_x x),sin(m\omega_x x)]
 \label{p2:fourier1d}
\end{equation}

In the same way, a neural network can incorporate 2D periodic constraints by utilizing a two-dimensional Fourier feature embedding \cite{wang2022respecting}.
\begin{equation}
 p(x,y)=\begin{bmatrix}
  cos(\omega_x x)cos(\omega_y y),...,cos(n\omega_x x)cos(m\omega_y y)\\
  cos(\omega_x x)sin(\omega_y y),...,cos(n\omega_x x)sin(m\omega_y y)\\
  sin(\omega_x x)cos(\omega_y y),...,sin(n\omega_x x)cos(m\omega_y y)\\
  sin(\omega_x x)sin(\omega_y y),...,sin(n\omega_x x)sin(m\omega_y y)\\
 \end{bmatrix}
 \label{p2:fourier2d}
\end{equation}

\noindent where $\omega_x=\frac{2\pi}{P_x}$, $\omega_y=\frac{2\pi}{P_y}$, $m,n$ are non-negative integers, $P_x$ and $P_y$ are the periods in the $x$ and $y$ directions. Dong et al. \cite{dong2021method} proved that any network representation such as $u_{NN}(v(x))$ or $u_{NN}(p(x,y))$ is periodic in the $x$ and $x,y$ directions, respectively.

\section{Results}
\label{p2:Results}
\subsection{Dynamic case: 1D Allen-Cahn equation}
\label{p2:1DACahn}
The Allen-Cahn equation is a compelling benchmark for PINNs due to its challenging characteristics. It is regarded as a ``stiff" PDE, a term that describes equations that require careful numerical handling to avoid instability in their solutions. This complexity arises from its ability to produce solutions with pronounced, sharp transitions in both spatial and temporal dimensions. The {general PDE form} for the 1D Allen-Cahn is given by:

\begin{equation}
{\frac{\partial u}{\partial t} = \epsilon\frac{\partial^2 u}{\partial x^2} + f(u)},
\end{equation}

\noindent {where $\epsilon > 0$ is the diffusion coefficient and $f(u)$ is a non-linear function of $u$. For this study, we set $\epsilon = 10^{-4}$ and $f(u) = -5u^3+5u$.} The initial and periodic boundary conditions are given by:

\begin{equation}
u(0, x) = x^2 \cos(\pi x), \quad \forall x \in [-1, 1]
\end{equation}
\begin{equation}
u(t, x + 1) = u(t, x - 1), \quad \forall t \geq 0 \quad \text{and} \quad x \in [-1, 1]
\end{equation}

\begin{figure}[t]
 \centering
 \includegraphics[width=0.95\textwidth]{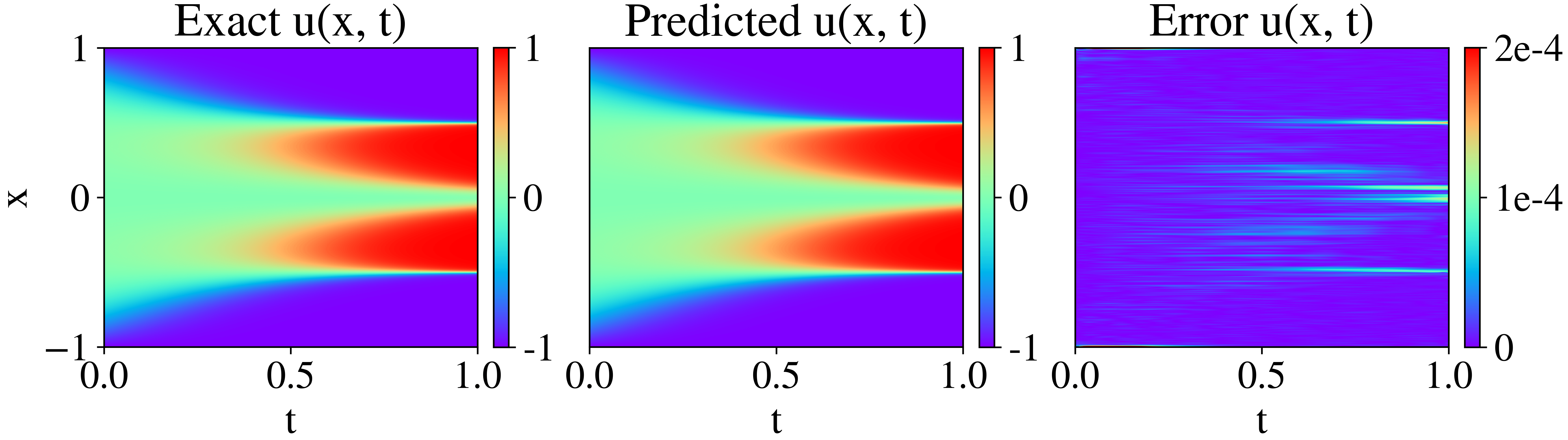}
 \caption[Solution of the 1D Allen-Cahn]{Exact solution of the 1D Allen-Cahn with the corresponding network prediction and the absolute error difference.}
 \label{p2:fig:AC}
\end{figure}

To obtain the solution of Allen-Cahn PDE in Fig. \ref{p2:fig:AC}, we train a PINN for $3\cdot10^5$ iterations with the standard Adam optimizer \cite{kingma2014adam} and an exponential learning rate scheduler on 25600 collocation points. For the periodic boundary conditions, we use the formulation of Section \ref{p2:ExactBCs} and set $w_{ic} = 100$, $w_{r} = 1$, following the benchmark setup of the recent literature \cite{wang2022respecting}. We also apply RBA only for the PDE residual term $\mathcal{L}_{r}$, while the rest of the parameters are mentioned in the appendix. The benchmark comparison of the final relative $L^2$ error obtained by different methods is presented in Table \ref{p2:t1}, where the error for the RBA weights represents the average of five runs with random seeds. It should be noted that only the best performing cited methods of Table \ref{p2:t1} (\cite{wang2022respecting}, \cite{mcclenny2020self}, \cite{zhang2023dasa}) utilize the PINN enhancements discussed in Section \ref{p2:enhancements}. Specifically for SA weights, we perform our own 5-run average based on \cite{mcclenny2020self} while also including our PINN enhancements for the sake of comparison. {For all examples}, the relative $L^2$ is defined as:

\begin{equation}
relative \ L^{2} = \frac{\lVert Exact - Predicted \rVert_{2}}{\lVert Exact \rVert_{2}}
\end{equation}

\begin{table}[H]
\centering
\caption[Relative $L^{2}$ comparison between different methods]{Relative $L^{2}$ comparison between different methods of the literature for the 1D Allen-Cahn equation. The RBA run represents the average of five randomly chosen seeds. {Note that the relative $L^2$ of SA weights is obtained by our own 5-run average based on \cite{mcclenny2020self}, including the enhancements of Section \ref{p2:enhancements} for fairness.}}
\begin{tabular}{cc}
\hline
Method & Relative $L^{2}$ error \\
\hline
Vanilla PINN \cite{raissi2019physics} & 4.98e-1 \\
Time marching \cite{mattey2022novel} & 1.68e-2 \\
Causal training \cite{wang2022respecting} & 1.39e-4 \\
DASA-PINNs \cite{zhang2023dasa}& 8.57e-5 \\
SA weights \cite{mcclenny2020self} & 1.11e-4 {$\pm$ 0.46e-5}\\
\textbf{RBA weights} & \textbf{5.8e-5} {$\pm$ 0.74e-6} \\
\hline
\end{tabular}

\label{p2:t1}
\end{table}

\subsubsection{Ablation Study for Allen-Cahn}
\begin{figure}[t]
  \centering
  \includegraphics[scale=0.625]{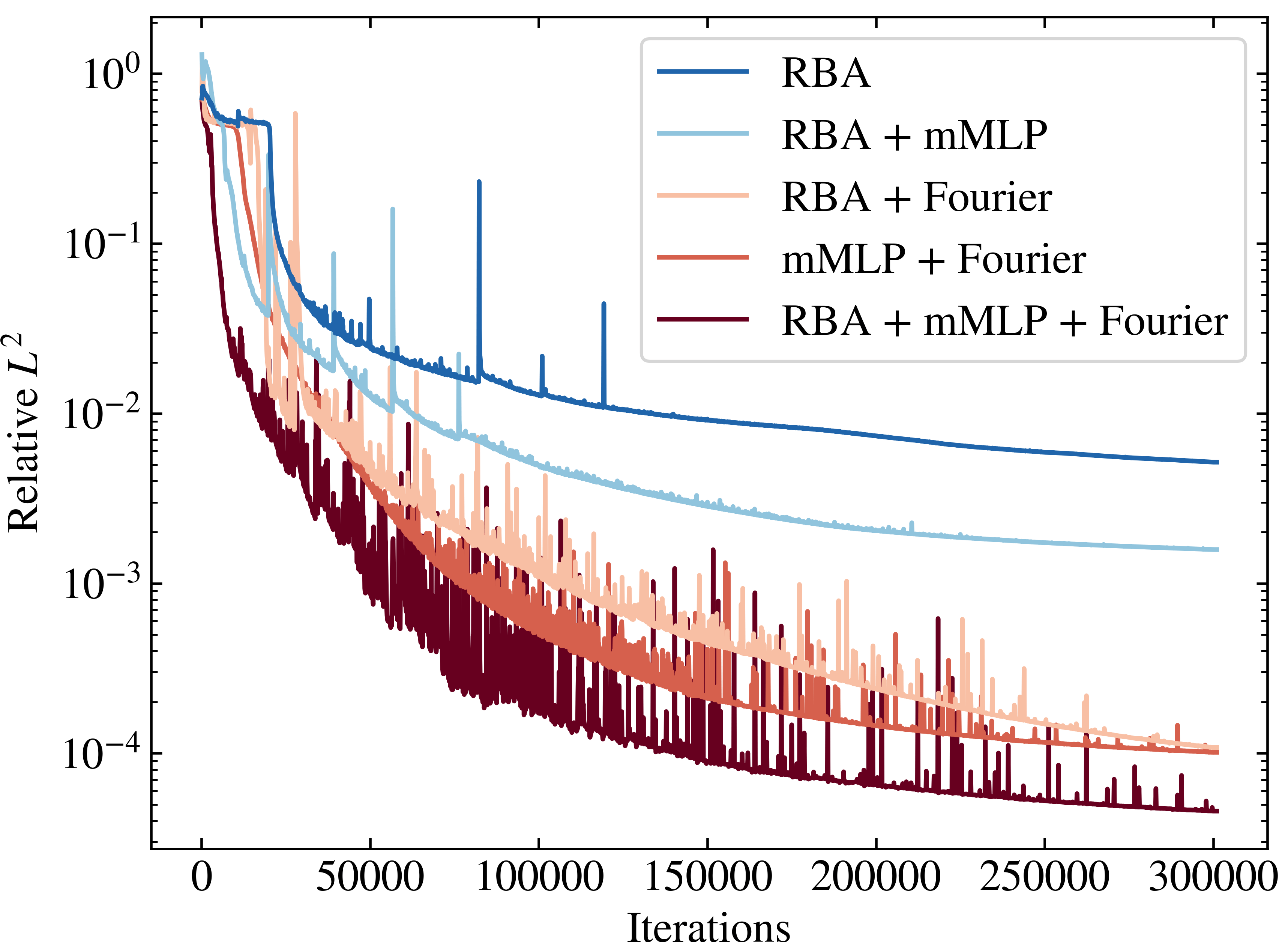}
  \caption[Ablation study convergence for the 1D Allen-Cahn]{\textbf{Ablation study convergence for the 1D Allen-Cahn}: Progression of convergence for each {component of the ensemble}. The results clearly demonstrate that the integration of the RBA approach and the Fourier feature embedding is crucial for attaining a minimal relative $L^2$. When combined with a modified MLP architecture, an optimal $L^2$ of $4.57\cdot 10^{-5}$ is reached. Additionally, the RBA weights address the instability issues in the standard PINN, achieving a relative $L^2$ of $5.2\cdot 10^{-3}$ without supplementary components. The noise linked with the top-performing methods suggests that the optimization process successfully avoids becoming trapped in sub-optimal solutions by effectively ``leaping" past local minima in the loss landscape. }
  \label{p2:fig:ablation_AC}
 \end{figure}

An ablation study involves measuring the performance of a system after removing one or more of its components to measure the relative contribution of the ablated components \cite{toscano2023teeth}.
To investigate the effect of {each component of the ensemble} on the relative $L^2$ convergence, we perform a single-component ablation study and report the results in Table \ref{p2:t2}. The best-performing seed of Table \ref{p2:t1} is selected to represent the full model. For each run, a component is removed to identify the effect of the omitted method on the performance, {as shown in the convergence histories of Fig. \ref{p2:fig:ablation_AC}}. As evident from these results, the Fourier feature embedding is the most important component to achieve low relative $L^2$. Moreover, when coupled with the RBA weights or the modified MLP, they reach a relative $L^2$ of $10^{-4}$. The full model leads to the best $L^2$ at $4.57\cdot 10^{-5}$. Moreover, the RBA weights initiate a steep convergence trajectory at 20000 iterations, achieving a final $L^2$ of $5.2\cdot 10^{-3}$ without any additional component. The noise associated with the best-performing methods indicates that the optimizer is ``jumping" over local minima of the loss landscape and effectively not getting stuck in a sub-optimal solution.

\begin{table}[H]
\centering
\caption[Final relative $L^2$ error for each case of the 1D Allen-Cahn]{Final relative $L^2$ error for each case of the single-component ablation study {for the ensemble} of the 1D Allen-Cahn.}
\begin{tabular}{cc}
\hline
Method & Relative $L^{2}$ error \\
\hline
RBA & 5.2e-3\\
RBA + mMLP & 1.58e-3 \\
RBA + Fourier & 1.0e-4\\
mMLP + Fourier& 1.0e-4\\
RBA + mMLP + Fourier & 4.57e-5\\
\hline
\end{tabular}
\label{p2:t2}
\end{table}
 
\subsection{Static case: 2D Helmholtz equation}
\label{p2:2DHMs}

The Helmholtz equation is commonly employed to model wave and diffusion phenomena, capturing changes over time in either a spatial or combined spatial-temporal domain. The 2D Helmholtz PDE is expressed as follows:

\begin{equation}
 \frac{\partial^2 u}{\partial x^2}+\frac{\partial^2 u}{\partial y^2}+k^2u-q(x,y)=0
 \label{p2:HM_eqn}
\end{equation}

\begin{equation}
\begin{split}
 q(x,y)=&-(a_1\pi)^2sin(a_1\pi x)sin(a_2\pi y)\\
   &-(a_2\pi)^2sin(a_1\pi x)sin(a_2\pi y)\\
   &+ksin(a_1\pi x)sin(a_2\pi y),
\end{split}
\label{p2:HM:FT}
\end{equation}

\noindent where $q(x,y)$ is the forcing term which leads to the analytical solution \\ $u(x,y)=sin(a_1\pi x)sin(a_2\pi y)$ \cite{mcclenny2020self}. The boundary conditions are defined as follows:

\begin{equation}
 u(-1,y)=u(1,y)=u(x,-1)=u(x,-1)=0
 \label{p2:HM:BC}
\end{equation}

To allow a direct comparison with previous studies, we set $a_1=1$, $a_2=4$, $k=1$ and follow the experimental setup described in \cite{zhang2023dasa}, setting $w_{bc} = 100$. We again apply RBA only for the PDE residuals $\mathcal{L}_{r}$. We initially apply a combination of Dirichlet and Periodic boundary conditions, with the latter enforced using the Fourier feature embedding. However, to avoid encoding the PDE analytical solution (i.e., $\sin(\pi x)\sin(4\pi y)$), we expand our inputs (i.e., $x,y$) using a truncated embedding with $m=n=5$ (Eq.~\ref{p2:truncfourier2d}):
\begin{equation}
 p(x,y)=\begin{bmatrix}
  cos(\omega_x x)cos(\omega_y y),...,cos(n\omega_x x)cos(n\omega_y y)\\
  cos(\omega_x x)sin(\omega_y y),...,cos(n\omega_x x)sin(n\omega_y y)\\
  sin(\omega_x x)cos(\omega_y y),...,sin(n\omega_x x)cos(n\omega_y y)\\
 \end{bmatrix}
 \label{p2:truncfourier2d}
\end{equation}

\begin{figure}[H]
 \centering
 \includegraphics[width=0.95\textwidth]{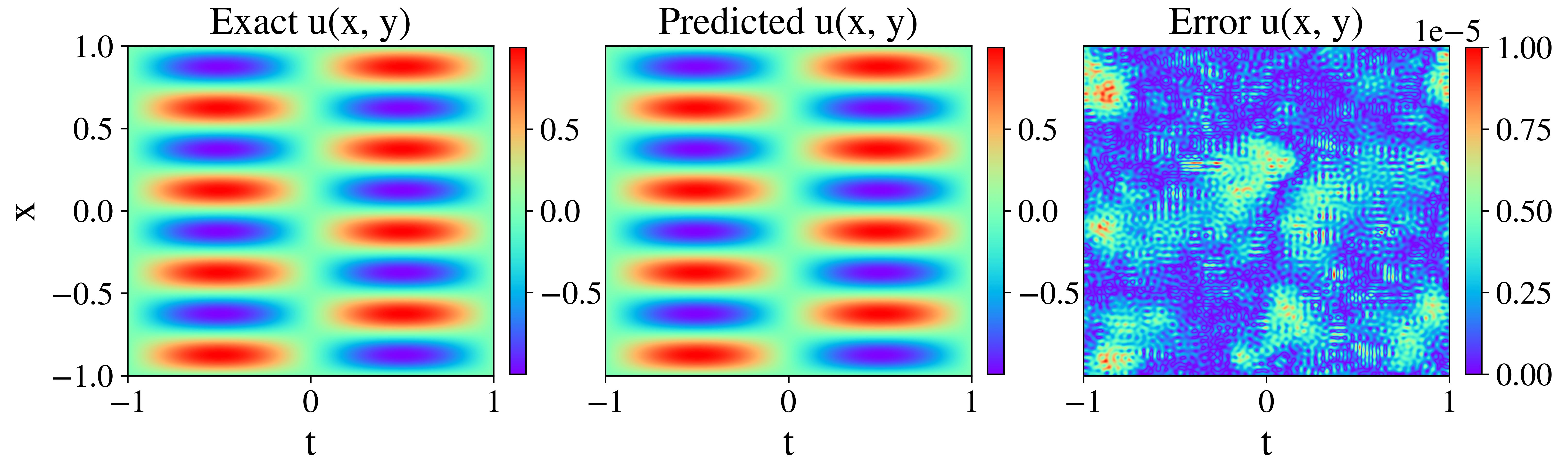}
 \caption[Solution for the 2D Helmholtz equation]{Analytical solution for the 2D Helmholtz equation and the corresponding network prediction with the absolute error difference.}
 \label{p2:fig:HM}
\end{figure}

Then, we approximate the solution of the PDE as $u_{F}=u_{NN}(p(x,y))$, where $u_{NN}$ is the modified MLP. Additionally, to demonstrate the flexibility of RBA weights, we examine an additional case applying only Dirichlet boundaries with approximate distance functions (ADF). Towards this end, we split the BCs into four partitions, i.e., $y=1$, $y=-1$, $x=1$, and $x=-1$, and use $\phi=(x^2-1)(y^2-1)$ as the composite distance function. Based on equation~\ref{p2:HM:BC} we define $g(\mathbf{x})=0$ and approximate the solution of the Helmholtz equation as:

\begin{equation}
u_{ADF}=(x^2-1)(y^2-1)u_{NN}(x,y),
\label{p2:HM_ADF}
\end{equation}

\noindent which exactly satisfies the BCs. We train our models (i.e., $u_{F}$ and $u_{ADF}$) using the Adam optimizer for $2\cdot 10^4$ iterations, followed by L-BFGS for $2\cdot 10^3$ iterations. {The exact solution of the 1x4 Helmholtz PDE, alongside the corresponding prediction obtained from the PINN is shown in Fig. \ref{p2:fig:HM}}. To allow a direct comparison with the self-adaptive weights (SA), we {also} train a second model with identical hyperparameters and update the local weights using gradient ascend as defined in \cite{mcclenny2020self}. {The average $L^2$ errors for the RBA and SA weights with Fourier and ADF are shown in Table \ref{p2:t3}}, along with the results reported by the current state-of-the-art methods.

\begin{table}[H]
\centering
\caption[Relative $L^{2}$ comparison between different methods of the literature for the 2D Helmholtz equation]{Relative $L^{2}$ comparison between different methods of the literature for the 2D Helmholtz equation. The RBA and SA values represent the average result of five randomly chosen seeds. {The relative $L^2$ of SA weights is obtained by our own 5-run average based on \cite{mcclenny2020self}, including the enhancements of Section \ref{p2:enhancements} for fairness.}}
\begin{tabular}{cc}
\hline
Method & Relative $L^{2}$ error \\
\hline
Vanilla PINN \cite{wang2021understanding}&8.14e-2 \\
Learning-rate annealing \cite{wang2021understanding} & 1.49e-3 \\
PECANN-AL \cite{basir2022investigating} & 2.77e-4 \\
AL-PINNs \cite{son2023enhanced}& 7.46e-4\\
DASA-PINNs \cite{zhang2023dasa} & 5.35e-5 \\
SA weights (ADF) \cite{mcclenny2020self} & 1.97e-4 {$\pm$ 4e-5} \\
SA weights (Fourier) \cite{mcclenny2020self} & 4.80e-5 {$\pm$ 9e-6} \\
RBA weights (ADF) &7.73e-5 {$\pm$ 1e-5}\\
\textbf{RBA weights (Fourier)} & \textbf{8.21e-6 {$\pm$ 3e-6}} \\
\hline
\end{tabular}

\label{p2:t3}
\end{table}

\subsubsection{Ablation Study for Helmholtz}

\begin{figure}[H]
 \centering
 \includegraphics[scale=0.625]{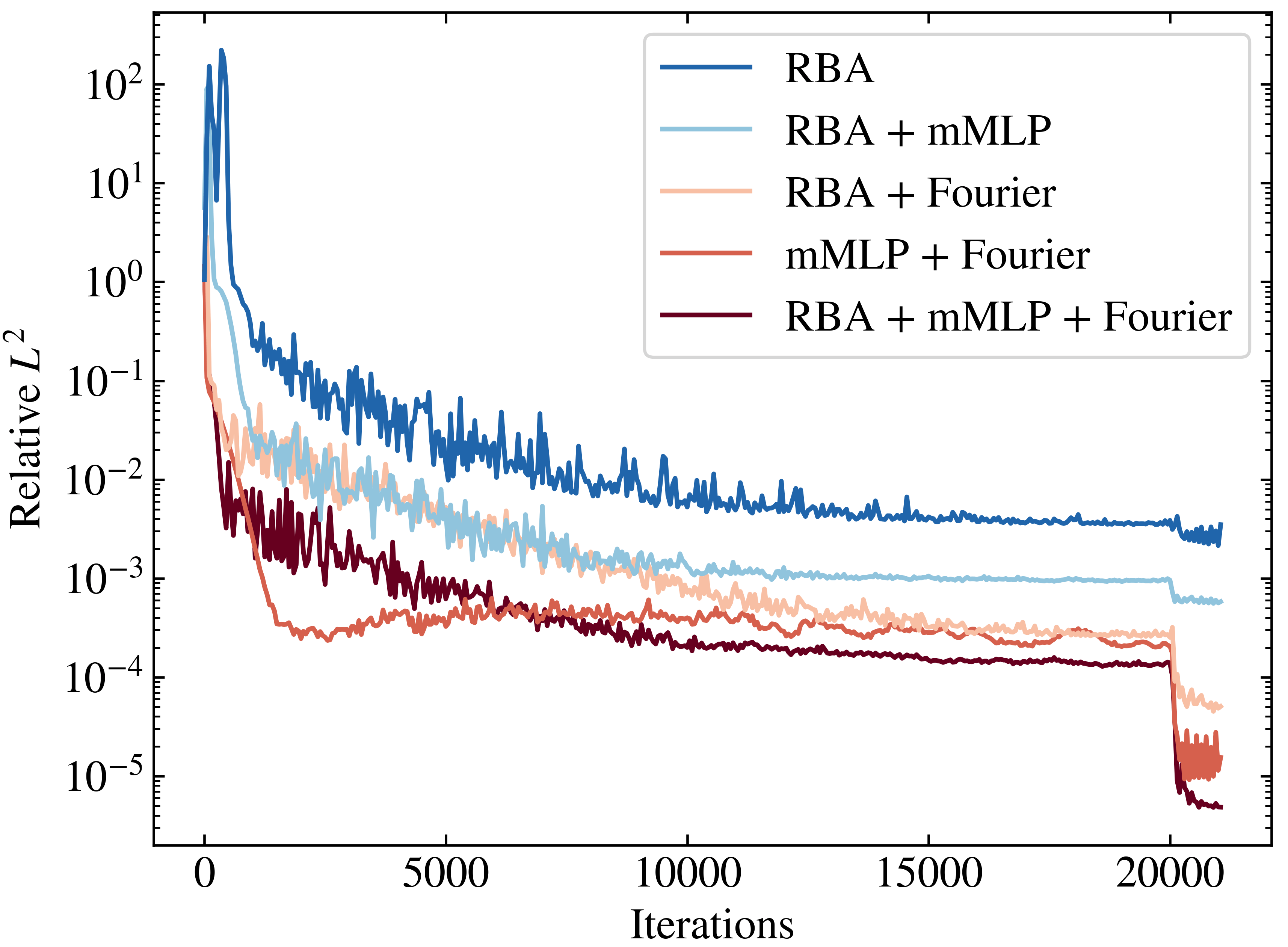}
 \caption[Ablation study convergence for the 2D Helmholtz]{\textbf{Ablation study convergence for the 2D Helmholtz}: The convergence trajectory for each experiment is illustrated here. For this case, the Fourier feature embedding is critical in achieving a low relative $L^2$. Coupled with the RBA weights and a modified MLP architecture, the model reaches an optimal $L^2$ of $5.18\cdot 10^{-6}$ after $2\cdot10^4$ Adam and $2\cdot10^3$ L-BFGS training steps.}

 \label{p2:fig:ablation_HM}
\end{figure}

In this ablation study for the 2D Helmholtz {equation}, we present the convergence histories for each ablated system (Fig. \ref{p2:fig:ablation_HM}) and summarize the final $L^2$ errors in Table \ref{p2:t4}. For the first case (i.e., $u_F$), the Fourier feature embedding has the most significant influence in reaching a minimized relative $L^2$. The accuracy further improves paired with either the RBA scheme or the mMLP, {while} the best-performing seed leads to an $L^2$ of $5.18\cdot 10^{-6}$; however, we note that this method is employed for comparison purposes since the Fourier features incorporate the trigonometric traits of the analytical solution. Therefore, we also perform the same study using distance functions ($u_{ADF}$). In general, exactly imposing the boundary conditions is important for accuracy, as it reduces the number of loss terms and allows the optimizer to focus on minimizing a single loss objective.

\begin{table}[H]
\centering
\caption[Relative $L^2$ error of the Helmholtz ablation study]{Final relative $L^2$ error for each case of the ablation study for the 2D Helmholtz {ensemble} using either the Fourier feature embeddings for exact periodic BC or ADF for exact Dirichlet BC.}
\begin{tabular}{cc}
\hline
Method & Relative $L^{2}$ error \\
\hline
RBA & 3.66e-3 \\
RBA + mMLP&5.91e-4 \\
\hline
mMLP + Fourier & 1.50e-5 \\
RBA + Fourier& 5.57e-5 \\
RBA + mMLP + Fourier & 5.18e-6 \\
\hline
mMLP + ADF & 6.34e-5 \\
RBA + ADF& 1.52e-3\\
RBA + mMLP + ADF & 5.96e-5 \\
\hline
\end{tabular}
\label{p2:t4}

\end{table}

\subsection{Inverse case: 3D Navier-Stokes}

To {further} investigate the effectiveness of our approach in real-world applications, we combine RBA with artificial intelligence velocimetry (AIV) to model the cerebrospinal fluid flow (CSF) in the perivascular spaces (PVS). Accurate quantification of CSF velocity and pressure is critical for understanding the glymphatic system (GS), which is related to the brain's waste removal and nutrient delivery. Building on the work from Boster et al., \cite{boster2023artificial}, we infer 3D high-resolution velocity pressure fields from 3D boundaries (BCs) and 2D plane velocities collected from the PVS of a murine's brain. {The data utilized in this study comes from the boundary conditions and internal flow velocity measurements. Specifically, the internal flow velocity measurements are obtained using particle tracking velocimetry (PTV), and the velocities of the boundary conditions are determined from the geometry, assuming non-slip boundary conditions.}

\begin{figure}[H]
 \centering
 \includegraphics[width=0.8\textwidth]{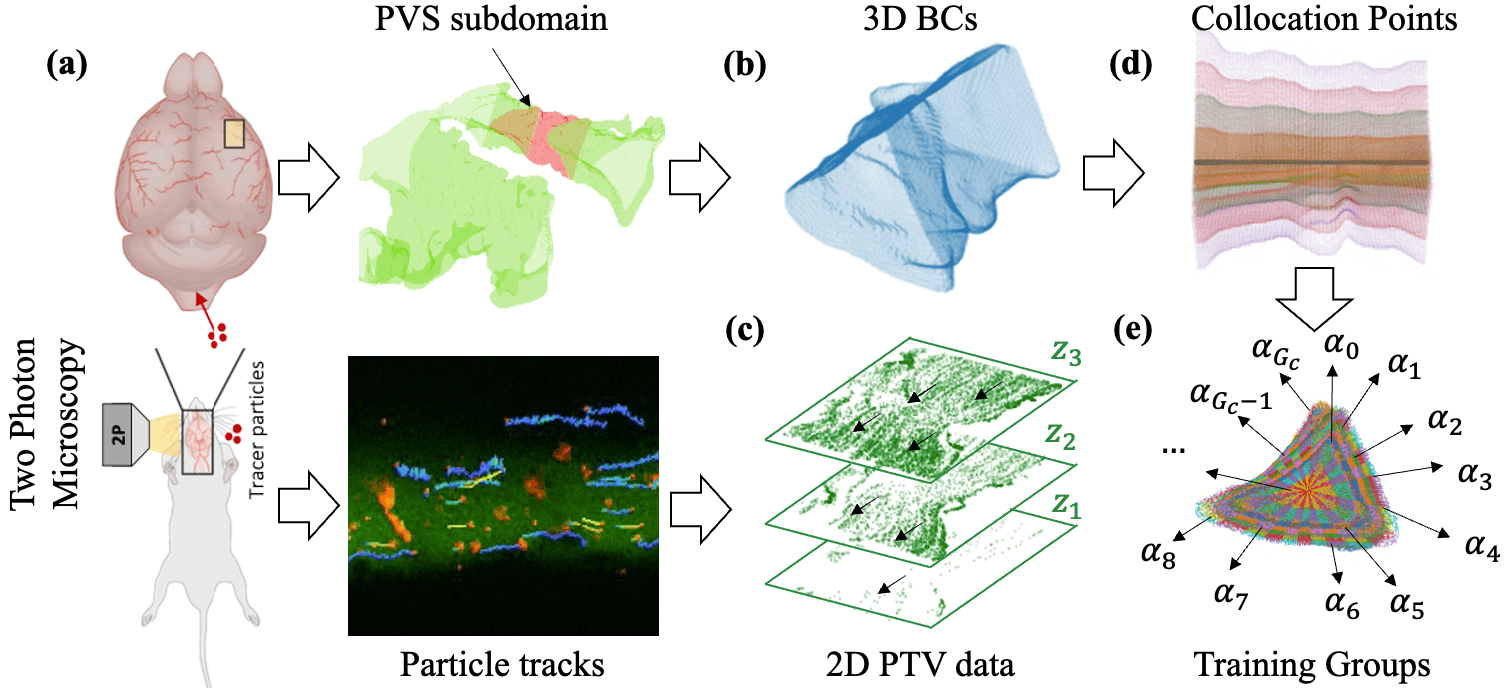}
     \caption[Artificial intelligence velocimetry data collection and generation]{\textbf{Artificial intelligence velocimetry data collection and generation}: (a) Tracer particles and one-micron microspheres are injected into a murine's brain to quantify and visualize the cerebrospinal flow in the perivascular spaces (PVS). The particle tracks are identified via two-photon microscopy, and their corresponding velocities are determined using particle tracking velocimetry. The PVS geometry is inferred from a volume scan. (b) 3D boundary conditions are specified from a PVS subdomain \cite{boster2023artificial}. (c) 2D velocity data is obtained from the particle tracks in three different planes. (d) The collocation points are generated by shrinking and combining the boundary conditions. (e) The collocation points are classified into $G_{c}$ training groups ($\alpha_i$) based on the spatial and temporal coordinates.}
 \label{p2:fig:data}
\end{figure}

To obtain the PTV data, Boster et al. \cite{boster2023artificial} injected one-micron microsphere tracers into the CSF, enabling the visualization and quantification of flow within the PVS. The microsphere motion was captured using PTV, and the PVS geometry was reconstructed from a 3D volume scan (see Fig.~\ref{p2:fig:data}(a)). Subsequently, \cite{boster2023artificial} identified a PVS subdomain from which they isolated the boundary conditions (BCs) and collected 2D velocity data points from three different planes, as illustrated in Fig.~\ref{p2:fig:data}(b) and (c). {Finally, to enhance the temporal resolution of the PTV dataset, the authors consolidated PTV data from multiple cardiac cycles into a single (phase-merged) cardiac cycle of $T=0.303 s$. In this study, to evaluate the PDE residuals during training, we generated collocation points by shrinking and combining the boundary conditions as shown in (Fig.~\ref{p2:fig:data}(d))}, which allowed us to control the spatial distribution of points within our domain.
\label{p2:3DNS}

\begin{figure}[H]
 \centering
 \includegraphics[width=0.82\textwidth]{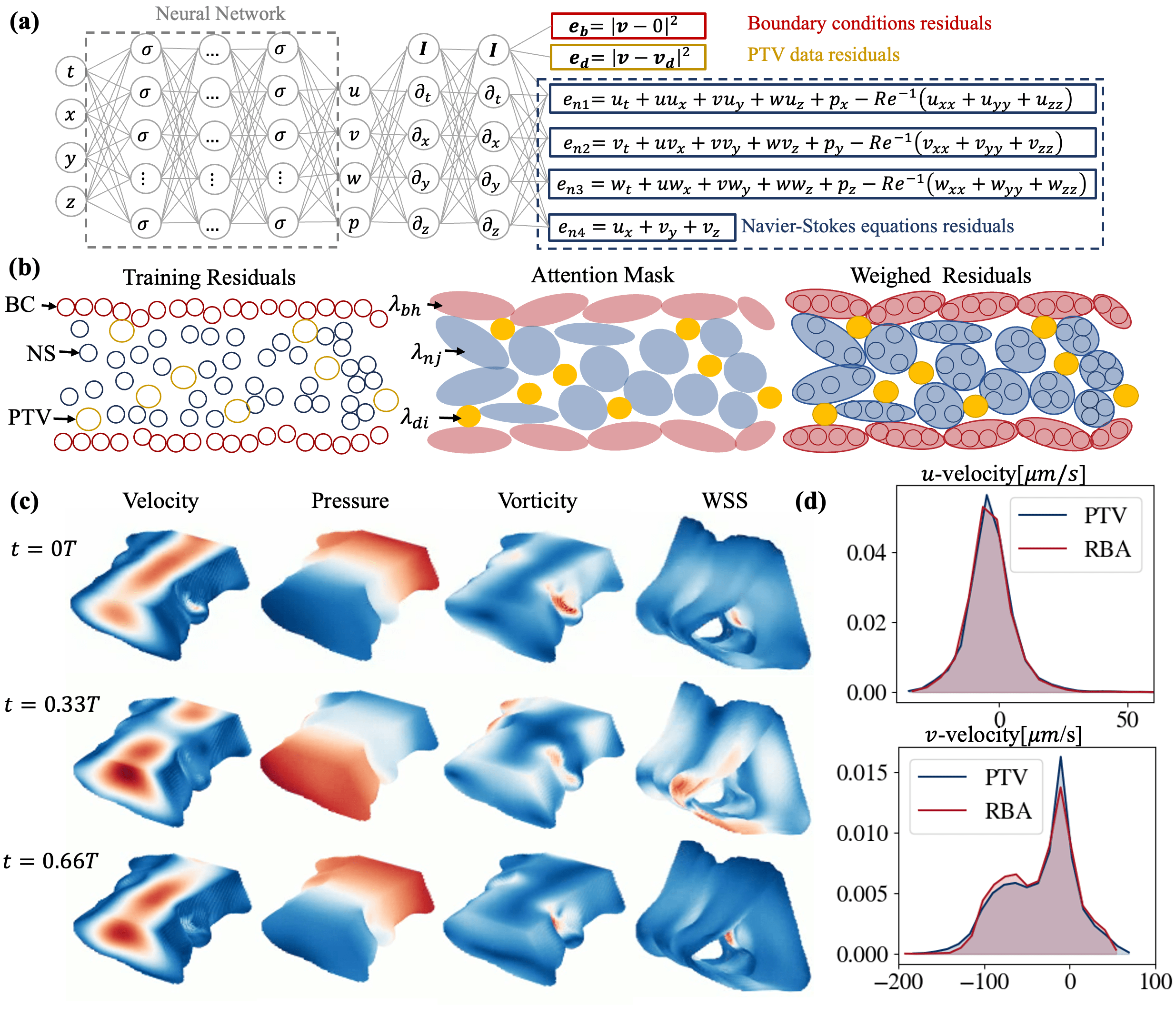}
 \caption[Artificial intelligence velocimetry framework]{\textbf{Artificial intelligence velocimetry framework:} (a) Schematic diagram of the AIV neural network. (b)  Each loss term is computed by element-wise multiplication between the training residuals and their corresponding RBA. The RBA is shared for the NS and boundary residuals. However, individual RBA weights are assigned for each data point. (c) Inferred 3D high-resolution velocity, pressure, vorticity magnitude, and wall shear stress (WSS) at $t=0T, 0.33T$, and $0.66T$. Notice that the proposed approach successfully captures the back-flow occurring at $t=0.33T$. (d) Histograms of the velocity components for the PTV validation dataset during the entire cardiac cycle ($T$).}
 \label{p2:fig:AIV_framework}
\end{figure}

 Generally, obtaining an attention mask requires processing at least as many multipliers as training points, which may hinder their optimal application for large datasets due to their computational cost. To address this issue, we propose using ordered mini-batches, which enable us to use a single attention mask for every minibatch. Specifically, we partition our collocation points and boundary conditions (BCs) into \(G_{c}\) and \(G_{b}\) training groups based on their spatial and temporal coordinates (see Fig.~\ref{p2:fig:data}(e)). Mini-batches are then created by selecting one point from each training group. Following this approach, we can share the same multiplier along the training group (Fig.~\ref{p2:fig:AIV_framework}(b)), reducing the computational cost and updating the attention mask at the same rate as the network parameters (i.e., per mini-batch). Given the large amount of collocation and BC points required to capture the 3D flow over $T$, we train our model using ordered mini-batches for the BCs and the system of PDEs while for the PTV data we use the entire dataset.

The governing physical laws for this problem are the Navier-Stokes (NS) equations, given by:

\begin{align}
  \frac{\partial\mathbf{v}}{\partial t} + (\mathbf{v}\cdot\mathbf{\nabla})\mathbf{v} &=-\mathbf{\nabla} p+\frac{1}{Re} \nabla^2\mathbf{v} \\
  \mathbf{\nabla}\cdot\mathbf{v}&=0\text{,}
\end{align} 

\noindent hence, following previous studies \cite{raissi2020hidden, jin2021nsfnets,boster2023artificial,zhang2023artificial}, we use a neural network to obtain a NS solution (i.e., 3D velocities and pressure) that follows the flow defined by the  PTV data and BCs (Fig.~\ref{p2:fig:AIV_framework}(a)). However, since the 3D velocities and pressure ($u,v,w,p$) have different scales, we redefine them using the scalars $(u_{s},v_{s},w_{s},p_{s})$  that ensure the network outputs have a similar order of magnitude. In particular, we approximate $u,v,w$ and $p$ as follows:

\begin{align}
    u&=u_su_{NN}(t,x,y,z)\\
    v&=v_sv_{NN}(t,x,y,z)\\
    w&=w_sw_{NN}(t,x,y,z)\\
    p&=p_sp_{NN}(t,x,y,z)\text{,}
\end{align}

\noindent where $t,x,y,z$ are the inputs (i.e., time and position) and $u_{NN},v_{NN},w_{NN},p_{NN}$ are the neural network outputs. Finally, to stabilize the training process and speed up convergence, we reparameterize our network using weight normalization (WN)\cite{salimans2016weight}. 

 During training, we apply global weights ($w_{\beta=\{b,d,n\}}$) to balance the contribution of the multiple loss terms, namely four NS (i.e., conservation of mass and conservation momentum in $x$, $y$, and $z$), two PTV data (i.e., velocity in $x$ and $y$), and three non-slip BCs (i.e., velocity in $x$, $y$ and $x$). Additionally, we apply RBA weights to control the interplay inside each loss term. In this step, we share RBA weights along the training groups for the residuals of the BC ($e_{bh}$) and NS ($e_{nj}$) while we assign individual multipliers for the PTV data ($e_{di}$) (Fig.~\ref{p2:fig:AIV_framework}(b)). We initialize the RBA with zero and update them using Eq.~\ref{p2:Update_RBA}. However, given the uncertainty in the BCs \cite{boster2023artificial}, we constrain them lightly by assigning them a lower RBA learning rate (i.e., $\eta^*$), which induces a smaller upper bound and within a smaller contribution. Finally, we define the total loss function as  $\mathcal{L}=\mathcal{L}_{bc}+\mathcal{L}_{d}+\mathcal{L}_{ns}$, where the combined loss terms for BCs ($\mathcal{L}_{bc}$), PTV data ($\mathcal{L}_{d}$) and NS ($\mathcal{L}_{ns}$) are defined as follows:
\begin{align}
  \mathcal{L}_{bc}& = \sum_{b=1}^{3}w_{b}\langle (\lambda_{h} \cdot e_{h,b})^2\rangle_h\\
  \mathcal{L}_{d}& = \sum_{d=1}^{2}w_d\langle(\lambda_{i} \cdot e_{i,d})^2\rangle_i\\
  \mathcal{L}_{ns}& = \sum_{n=1}^{4}w_{n}\langle(\lambda_{j} \cdot e_{j,n})^2\rangle_j\text{,}
\end{align}

\noindent here, $e_{h,b}, e_{i,d}, e_{j,n}$ are the BC, data and NS residuals for points $h,i$, and $j$ (see Fig.~\ref{p2:fig:AIV_framework}(a)), $w_{b}, w_d, w_{n}$ are global weights applied to each loss term, $\lambda_{h}$, $\lambda_{j}$ are the RBA assigned for the BCs and NS training groups and $\lambda_{i}$ are the RBA assigned for each PTV point. We note that the BCs and NS are averaged over their corresponding batch sizes, while the data terms are averaged over the PTV points.

To evaluate the AIV performance, we split our PTV data into 60\% for training and 40\% for validation and train the model for  $2.5\cdot10^5$ iterations using Adam optimizer \cite{kingma2014adam}. Additional implementation details are described in ~\ref{p2:AIV_implementation}. Once the model is trained, we get a continuous and differentiable function for the velocity ($\mathbf{v}=[u,v,w]$) and pressure ($p$). Thus, following \cite{boster2023artificial} we use automatic differentiation to compute the vorticity ($\boldsymbol{\omega}$) and shear stress ($\tau$) as follows,
\begin{align}
    \boldsymbol{\omega}&=\boldsymbol{\nabla}\times \mathbf{v}\\
    \tau&=\mu\dot{\gamma}\text{ ,}
\end{align}
\noindent where $\mu$ is the viscosity, and $\dot{\gamma}$ is the shear rate computed from the stress-strain rate tensor ($\mathbf{E}$) described as,

\begin{align}
    \dot{\gamma}&=\sqrt{2\mathbf{E}:\mathbf{E}}\\
    \mathbf{E}&=\frac{1}{2}(\boldsymbol{\nabla}\mathbf{v}+(\boldsymbol{\nabla}\mathbf{v})^T)
\end{align}

As shown in Fig.~\ref{p2:fig:AIV_framework}(c), the proposed method can reconstruct 3D high-resolution velocity, pressure, vorticity, and wall shear stress (WSS) fields from 3D BCs and 2D plane velocities. Notice that our model successfully captures the backflow occurring at $t=0.33T$. Similarly, it can be observed that the irregular regions in the BCs induce high vorticity and  WSS. Finally, Fig.~\ref{p2:fig:AIV_framework}(d) shows that the predicted 2D velocities (i.e., $u$ and $v$) follow the same distribution as the PTV data from the validation dataset. Additional results for different time-steps along the cardiac cycle $T$ are shown in Fig.~\ref{p2:fig:Snap_AIV}.

To allow a direct comparison with the self-adaptive weights (SA), we train a second model with identical hyperparameters and update the local weights using gradient ascend as defined in \cite{mcclenny2020self}. Table~\ref{p2:tAIV} shows the relative $L^2$ error in the validation dataset for the 2D PTV velocity $(u,v$) in $x$ and $y$ directions. For reference, we also include the results obtained by \cite{boster2023artificial}, where the authors trained their model using Learning-rate annealing \cite{wang2021understanding} for $5\cdot10^5$ iterations (i.e., twice as many as us). Notice that applying global weights along with local multipliers is essential to achieve a low relative $L^2$ error, especially for the velocity in the $x$ direction.

\begin{table}[H]
\centering
\caption[Relative $L^{2}$ comparison for the validation dataset \cite{boster2023artificial}]{Relative $L^{2}$ comparison for the validation dataset between the original study \cite{boster2023artificial} and our approach using vanilla PINN and combining weight normalization with SA or RBA weights. The weighted PINN, RBA, and SA values represent the average result of five randomly chosen seeds.}
\begin{tabular}{ccc}
\hline
Method & Relative $L^{2}(u)$&Relative $L^{2}(v)$\\
\hline
Learning-rate annealing \cite{boster2023artificial}& 4.21e-1 & 2.11e-1\\
PINN + global weights & 3.31e-1 & 2.67e-1\\
SA weights \cite{mcclenny2020self} & 2.94e-1 {$\pm$ 2.5e-3} & 2.48e-1 {$\pm$ 2.3e-3} \\
\textbf{RBA weights} & \textbf{2.43e-1 {$\pm$ 1.6e-3}}& \textbf{2.10e-1 {$\pm$ 1.5e-3}}\\

\hline
\end{tabular}
\label{p2:tAIV}
\end{table}

\subsubsection{Ablation Study for AIV}

\begin{figure}[H]
 \centering
 \includegraphics[width=0.9\textwidth]{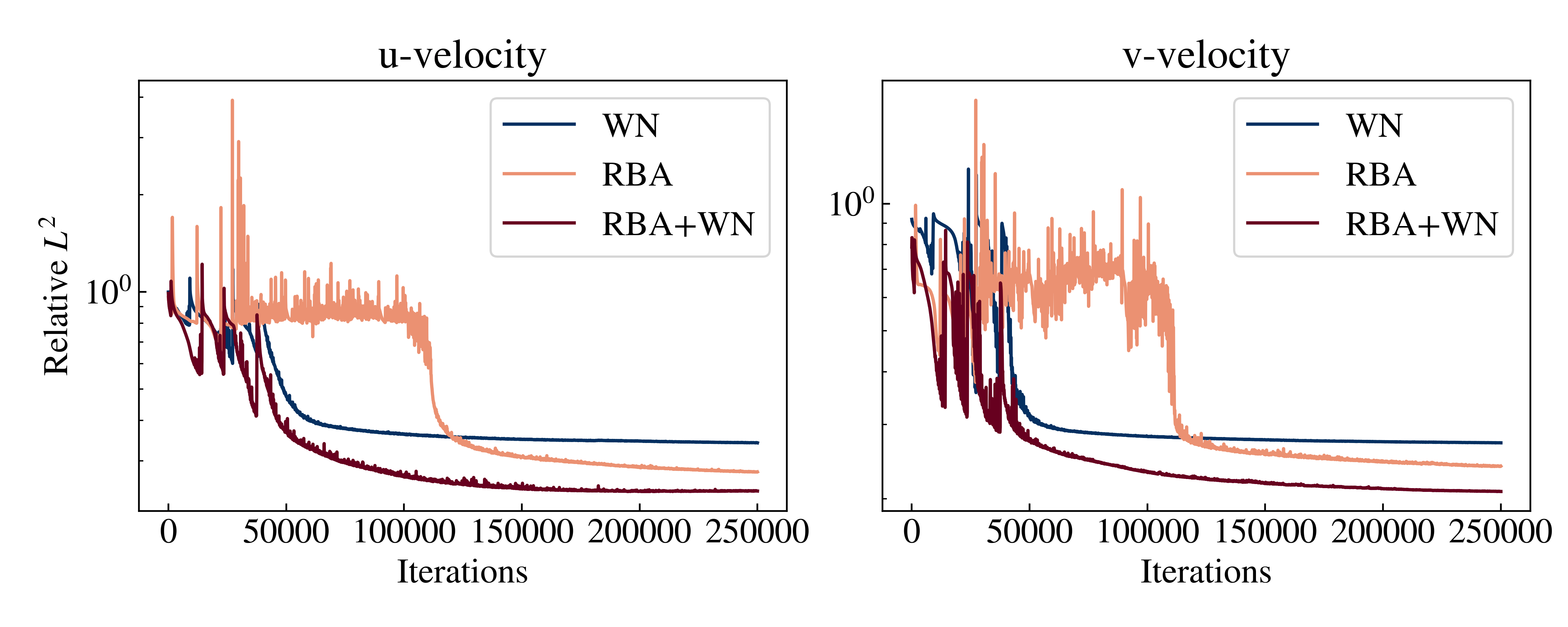}
 \caption[Ablation study convergence for AIV]{\textbf{Ablation study convergence for AIV}:  Weight normalization speeds up the model convergence. However, applying RBA multipliers is critical to achieving a low relative $L^2$. The full model, which combines RBA with weight normalization, reaches an optimal $L^2$ of $2.42\cdot 10^{-1}$ and $2.08\cdot 10^{-1}$ for the velocity in $x$ and $y$ (i.e., $u$ and $v$) after $2.5\cdot10^4$ Adam training iterations.}
 \label{p2:fig:ablation_AIV}
\end{figure}

The final relative $L^2$ errors for each ablated system on the validation dataset are summarized in Table \ref{p2:tAIV_AS}. As shown in Fig. \ref{p2:fig:ablation_AIV}, the RBA weights are crucial to achieving a low relative $L^2$ error; however, the WN is essential for accelerating convergence and stabilizing the training process. By combining RBA with WN, the full model achieves an optimal $L^2$ of $2.42 \cdot 10^{-1}$ and $2.08 \cdot 10^{-1}$ for the $u$ and $v$ velocity, respectively.

\begin{table}[H]
\centering
\caption[Final relative $L^2$ error for each case of the ablation study on the AIV]{Final relative $L^2$ error for each case of the ablation study on the AIV for the $u,v$ velocity using the validation dataset.}
\begin{tabular}{ccc}
\hline
Method & Relative $L^{2}(u)$&Relative $L^{2}(v)$\\
\hline
WN  & 3.41e-1 & 2.71e-1\\
RBA  & 2.77e-1 & 2.39e-1\\
RBA + WN& 2.42e-1 & 2.08e-1\\

\hline
\end{tabular}

\label{p2:tAIV_AS}
\end{table}

\subsection{{Comparison of RBA, SA weights and vanilla}}
\label{p2:SA}
{
Both RBA and SA weights are based on the same principle: assign higher weight to more important collocation points and increase the attention of the optimizer on certain regions of the domain. The key difference between the two methods is that SA weights accumulate the gradient term $\frac{\partial \mathcal{L}} {\partial \lambda}$ (through gradient ascent), while RBA weights accumulate the normalized residual $\frac{\left| e_i \right|}{\;\;\;\lVert e \rVert_{\infty}}$. Thus, RBA does not train any auxiliary $\lambda$, instead it directly uses a moving average of the residuals. In this section, we compare the performance of plain RBA with plain SA, against that of the vanilla PINN. Thus, we set the loss term weights $w$ to 1 for all cases and for RBA we set $\eta^*=0.001=1-\gamma$. For simplicity, we use the same collocation point sampling for Helmholtz as we do for Allen-Cahn and we also drop the periodic Fourier features. The rest of the settings remain the same as in the ablation studies. Note that for SA, we initialize $\lambda_{ic}$ with (0-100) for Allen-Cahn and $\lambda_{bc}$ with (0-1) for Helmholtz \cite{mcclenny2020self}.}

\begin{figure}[H]
 \centering
 \includegraphics[width=0.95\textwidth]{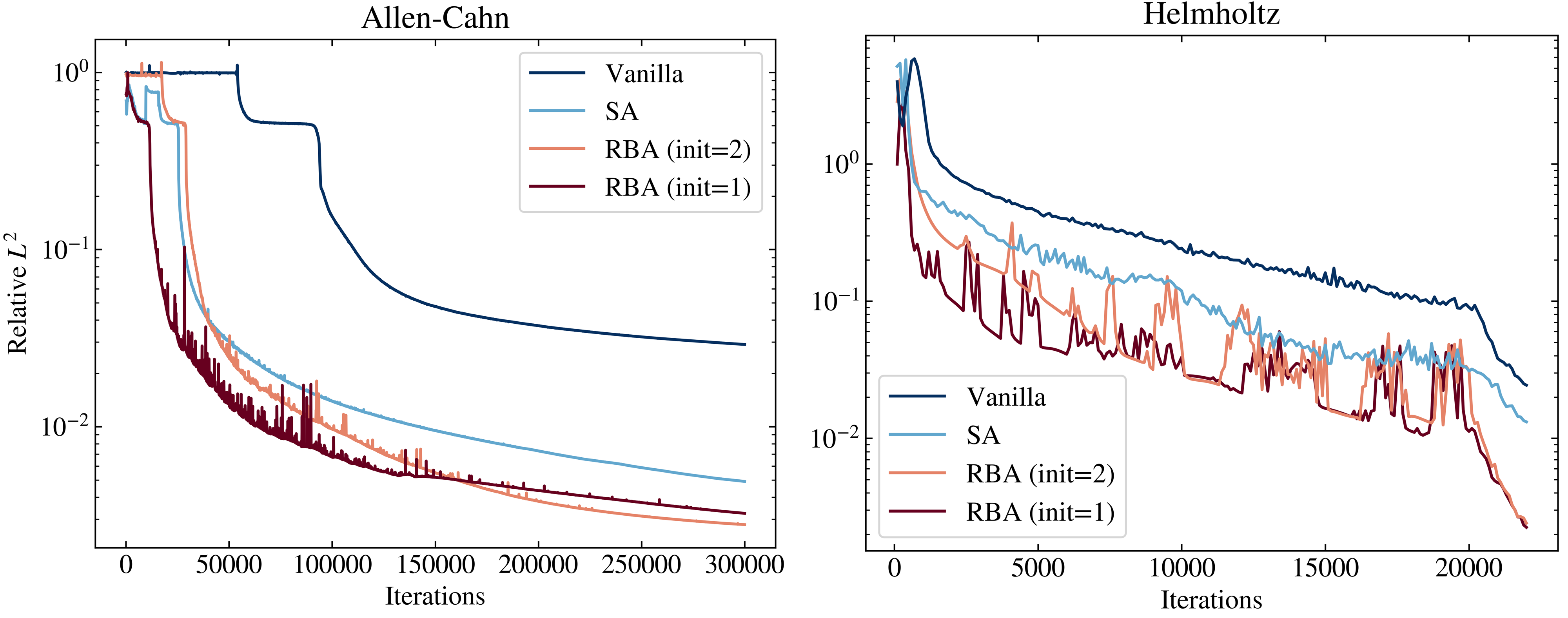}
 \caption[Convergence for plain RBA, plain SA and vanilla]{\textbf{Convergence for plain RBA, plain SA and vanilla}. Allen-Cahn (left) and Helmholtz (right). The RBA method achieves one order of magnitude faster convergence, even without initial/boundary term weights $w$. The SA method is on average four times faster than vanilla.}
 \label{p2:fig:RBA_conv}
\end{figure}

The convergence plots of Fig. \ref{p2:fig:RBA_conv} show that RBA accelerates the convergence of vanilla PINN by one order of magnitude. Note that for Allen-Cahn, the vanilla model is able to converge due to the learning rate scheduler. The SA weights also provide a substantial improvement to the vanilla, which is around four times faster for the given setup. The two initialization options of Algorithm 1  show similar performance. This indicates that the balance between the residuals is what leads to convergence and not a higher $w$ weight for the initial/boundary loss terms.

\begin{figure}[H]
 \centering
 \includegraphics[width=\textwidth]{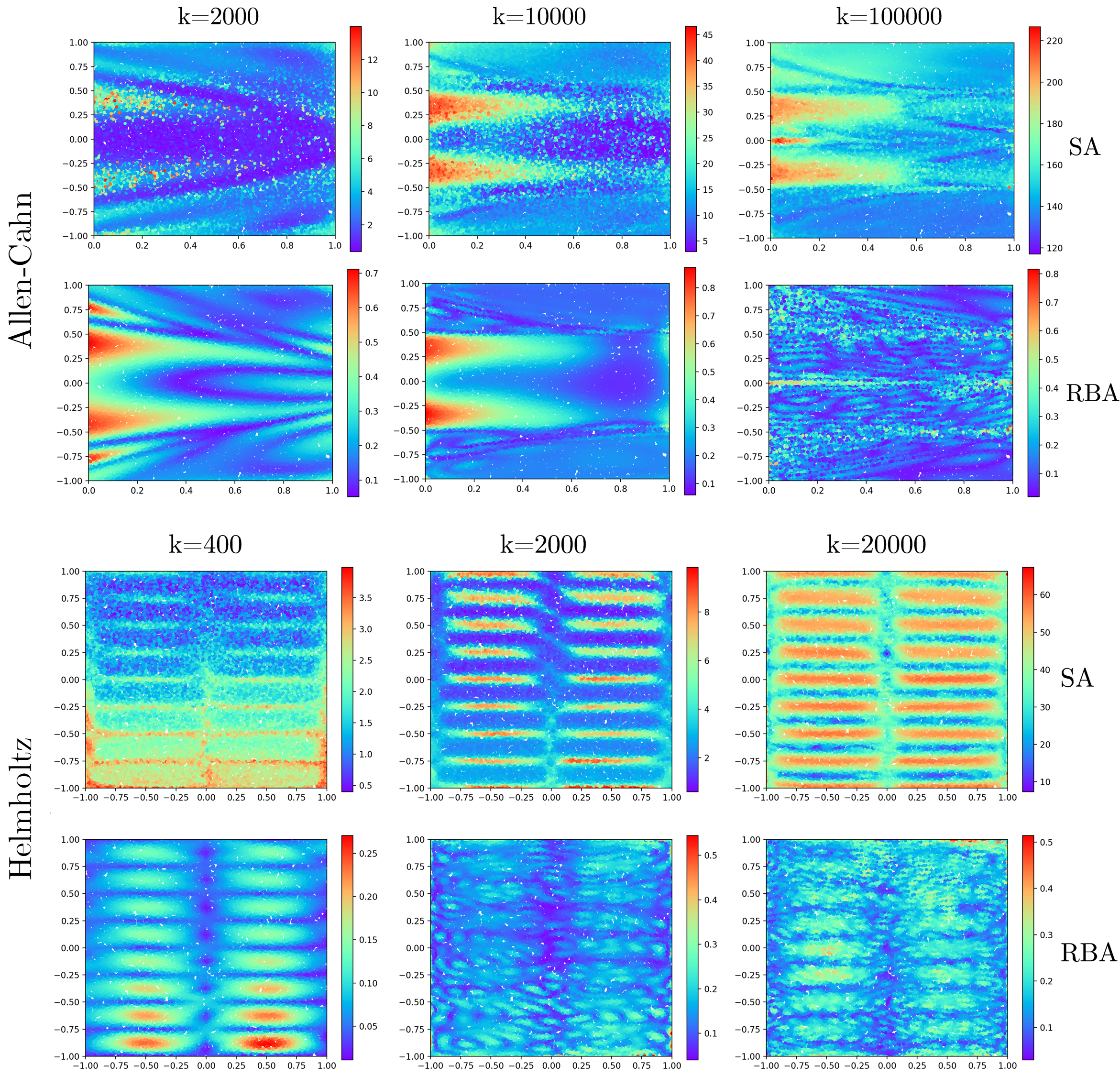}
 \caption[Weight maps for RBA and SA]{{\textbf{Weight maps for RBA and SA}. Both cases show relative agreement. However, RBA appear to produce a less noisy attention mask, which is more responsive in the initial iterations. Moreover, during the final iterations RBA becomes more homogeneous, which implies uniform domain convergence.}}
 \label{p2:fig:HM_SA}
\end{figure}

{
Additionally, both benchmarks of Fig. \ref{p2:fig:HM_SA}, show relative agreement in the regions of higher attention. However, RBA appear to produce a less noisy attention mask, which is more responsive in the initial iterations. Moreover, during the final iterations RBA becomes more homogeneous, which implies uniform domain convergence. Finally, while SA weights solve a min-max optimization problem (minimize the residuals while maximizing $\lambda$), the normalization with $\lVert e \rVert_{\infty}$ ensures that RBA weights are bounded (in this case from 0 to 1).}

{These results indicate that adaptive weight strategies like RBA and SA, have a significant impact on the vanilla PINN performance. Moreover, even in cases where we used ten times less collocation points, both RBA and SA converged to a reasonable solution, while the vanilla failed to do so. Therefore, we would like to emphasize the importance of local attention mechanisms, which effectively guide the optimizer by focusing on the essential information of the solution space.}

\section{Summary}
\label{p2:Summary}
 In this work, we propose a fast, residual-based attention {scheme} to enhance the accuracy of PINNs in both forward (i.e., static/dynamic) and inverse problems, competing with the state-of-the-art performing models of the literature. Additionally, we believe that an equally important element of {local-attention strategies} is in the realms of convergence. For example, the 1D Allen-Cahn equation is a hard problem to be solved by the vanilla PINN due to its stiff nature, but RBA can initiate an early convergence trajectory, without any empirical loss term weights $w$. Furthermore, {when combined with other PINN enhancements, the ablation study points out that an important aspect of achieving a solution accuracy of $\mathcal{O}(10^{-5})$} is the formulation of exact boundary conditions. The versatility of the proposed method allows for its application with a few additional lines of code and can be extended to complex problems with multiple loss terms. Therefore, we also combine it with artificial intelligence velocimetry (AIV) to model the cerebrospinal fluid flow (CSF) in the perivascular spaces (PVS). Our observations indicate that the RBA weights are critical to reduce the relative $L^2$ error and speed up convergence. Integrating the proposed attention scheme with existing methodologies can enable more reliable and accurate PINN or neural operator solutions in complex physical modeling scenarios.

\section*{Acknowledgements}

SJA and NS thank the Swiss National Science Foundation grant ``Hemodynamics of Physiological Aging" (Grant nr $205321\_197234$). JDT and GEK acknowledge support by the DOE SEA-CROGS project (DE-SC0023191), the MURI-AFOSR FA9550-20-1-0358 project, and the ONR Vannevar Bush Faculty Fellowship (N00014-22-1-2795). Finally, we thank Prof. N. Sukumar from the University of California, Davis, for engaging in insightful discussions on the approximate distance functions.

\appendix 

\section{Implementation Details}
\label{p2:Implementation Details}

\subsection{RBA weight evolution}

\begin{figure}[H]
 \centering
 \includegraphics[width=\textwidth]{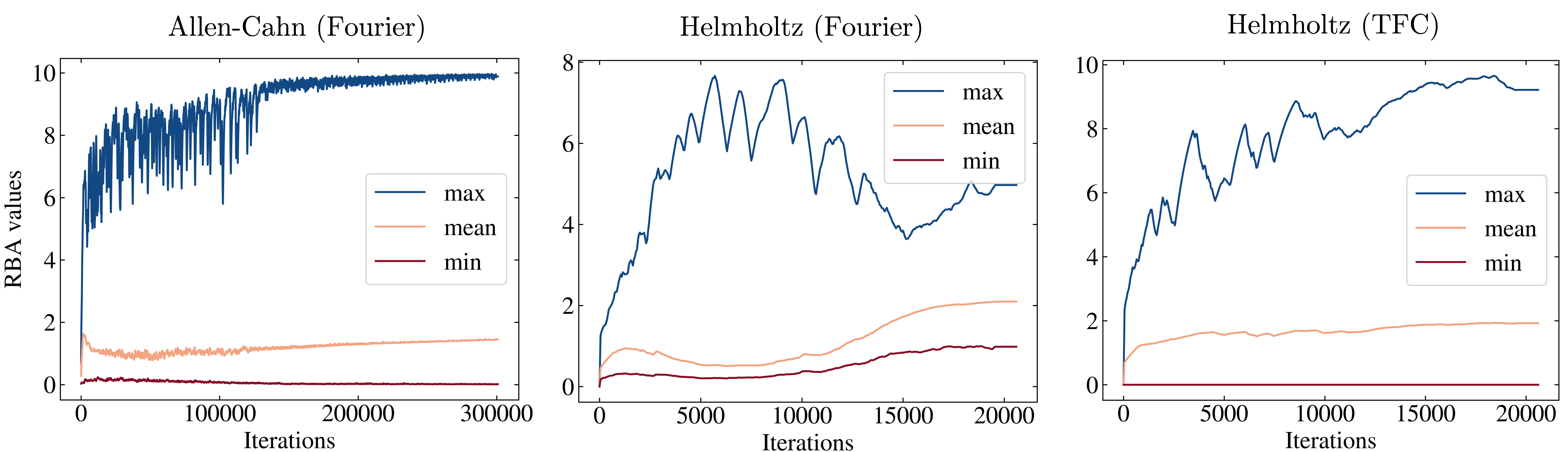}
 \caption[Evolution of RBA weights for Allen Cahn and Helmholtz]{\textbf{Evolution of RBA weights for Allen Cahn and Helmholtz}. For each case, the peak value is bounded by Eq. \ref{p2:eq:bound} and equals to $10$. Notice that the maxima fluctuate until they approach the upper limit during the final learning stages. On the other hand, the mean values show a less noticeable fluctuation before eventually stabilizing at about $\approx 20\%$ of the upper bound. This trend indicates that, on average, the total magnitude of weights remains constant while their distribution adjusts dynamically as the optimizer shifts its focus across different domain sections.}
 \label{p2:fig:lambdas}
\end{figure}

This section examines the evolution of the RBA values for the full model of the ablation study for each case. Figure \ref{p2:fig:lambdas} shows the evolution of RBA weights for Allen Cahn and Helmholtz {equations}. For each run, the maximum value is bounded by Eq. \ref{p2:eq:bound}, which for the ablation studies of Allen-Cahn and Helmholtz equals $10$. The distribution of weights significantly varies depending on the stage of training; hence, the maximum values fluctuate while approaching the upper bound. The fluctuation of the mean values is less pronounced until they finally converge to $\approx 20\%$ of the upper bound. This behavior indicates that, on average, the total magnitude of weights remains constant while their distribution varies as the optimizer focuses on different parts of the domain. Thus, by introducing the decay factor $\gamma$, we can effectively prevent exploding weight values. At the same time, this allows the weights to be flexibly distributed, adapting as needed throughout different stages of the solution process. 

\begin{figure}[H]
 \centering
 \includegraphics[width=\textwidth]{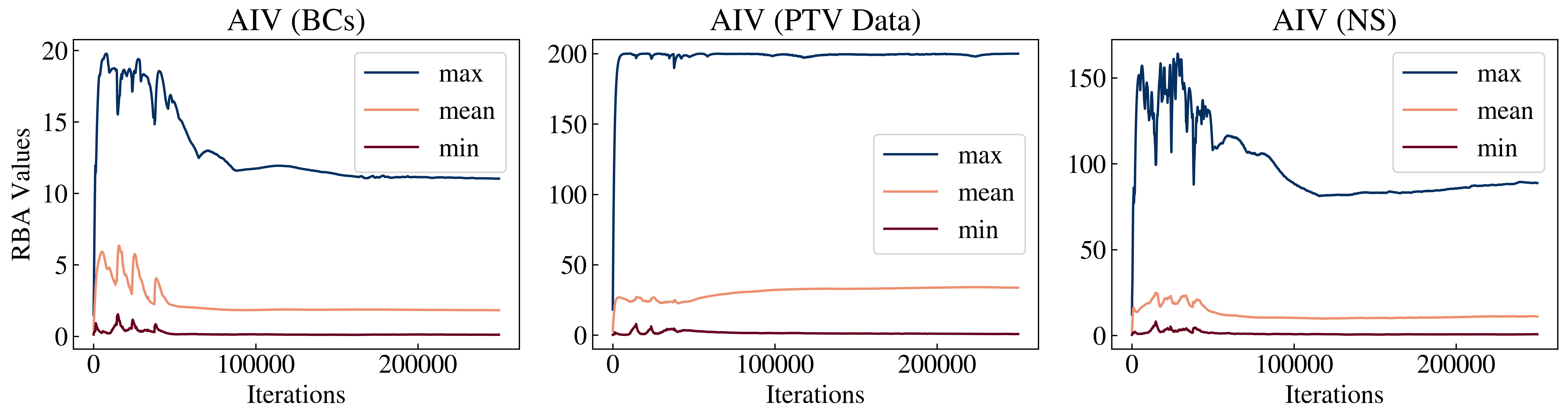}
 \caption[Evolution of RBA weights for AIV]{\textbf{Evolution of RBA weights for AIV}.  We illustrate the RBA multipliers' variation throughout the AIV training process. Although there are nine sets of RBAs (i.e., one for each loss term), we compute each group's maximum, mean, and minimum values (i.e., BCs, PTV data, and NS) to simplify visualization. Notice that the lower learning rate for the BCs leads to a lower upper limit. Also, the maximum RBA for the PTV data increases to its upper limit and remains constant along the training process, which suggests that there are unlearnable points in the dataset (i.e., outliers or contradictory information). On the other hand, the BCs and NS weights reach their max in the initial iterations and then start decreasing, indicating that the difficult points have already been learned. 
}
 \label{p2:fig:AIV_lambdas}
\end{figure}

Similarly, Figure \ref{p2:fig:AIV_lambdas} shows the variation of the RBA multipliers along the training process for the AIV. For this case, we have nine sets of RBAs (i.e., one per loss term); however, to ease the visualization, we compute the respective metrics (i.e., max, mean, and min) for each loss term group (i.e., BCs, PTV data, and NS). Notice that the larger learning rate ($\eta^*$) induces a higher upper bound. Also, the maximum value for the PTV data grows rapidly and remains constant along the training process. This may indicate that, since we are dealing with real data, there may be outliers or contradictory information that the model cannot learn. On the other hand, the BCs and NS RBA weights reach their maximum value in the first iterations and then start decreasing, indicating that the difficult regions have already been learned.

\subsection{Benchmark settings}

The network parameters are initialized using Xavier normal distribution \cite{glorot2010understanding}. The rest of the case-specific information and their corresponding training time are tabulated below (see Tables \ref{p2:AC_t4} - \ref{p2:tAIVspeed}). Note that the reported training time should only be used as an inner reference since the codes were not optimized for speed.

\begin{table}[H]
\centering
\caption[Allen-Cahn Implementation Details]{Allen-Cahn Implementation Details}
\begin{tabular}{ccc}
\hline
 \multicolumn{2}{c}{Implementation Details} \\\hline
Sampling method & Latin Hypercube\\
Number of collocation points & 25600 \\
Number of initial condition points & 512 \\
Number of hidden layers & 6\\
Neurons per layer & 128\\
Activation function& tanh\\
Reparametrization &mMLP\\
Initial learning rate & 1e-3 \\
Decay rate & 0.9\\
Decay step & 5000\\
RBA decay parameter ($\gamma$)&0.999\\
RBA learning rate ($\eta^*$)& 1e-2 \\
Offset ($\lambda_{o})$ & 0\\
ML framework & PyTorch \\
GPU &GeForce RTX $3090$ \\\hline

\end{tabular}

\label{p2:AC_t4}
\end{table}

\begin{table}[H]
\centering
\caption[Comparison of computational time for different combinations of the ablation study for the 1D Allen-Cahn]{Comparison of computational time for different combinations of the ablation study for the 1D Allen-Cahn.}
\begin{tabular}{
    >{\centering}m{5cm}
    >{\centering\arraybackslash}m{2.5cm}}
\hline
Method & Training time (ms/iteration) \\
\hline
Vanilla & 17 \\
RBA & 17 \\
RBA + mMLP & 58 \\
RBA + Fourier& 18\\
mMLP + Fourier & 58\\
RBA + mMLP + Fourier & 58\\
\hline
\end{tabular}

\label{p2:t1speed}
\end{table}

\begin{table}[H]
\centering
\caption{Helmholtz (Fourier) ($a_1=1$, $a_2=4$) Implementation Details}
\begin{tabular}{cc}
\hline
 \multicolumn{2}{c}{Implementation Details} \\\hline
Sampling method & Uniform grid\\
Number of collocation points & 1201\\
Number of boundary points & 200 \\
Number of hidden layers & 6\\
Neurons per layer & 128\\
Activation function& tanh\\
Reparametrization &mMLP\\
Initial learning rate & 5e -3 \\
Decay rate &0.7 \\
Decay step &1000 \\
RBA decay parameter ($\gamma$)&0.9999\\
RBA learning rate ($\eta^*$)& 1e-3 \\
Offset ($\lambda_{o})$ & 0\\
ML framework & Jax \\
GPU &GeForce RTX $3090$ \\\hline
\end{tabular}
\label{p2:H1x4_t4}
\end{table}

\begin{table}[H]
\centering
\caption{Helmholtz (ADF) ($a_1=1$, $a_2=4$) Implementation Details}
\begin{tabular}{cc}
\hline
 \multicolumn{2}{c}{Implementation Details} \\\hline
Sampling method & Uniform grid\\
Number of collocation points & 1201\\
Number of hidden layers & 6\\
Neurons per layer & 128\\
Activation function& tanh\\
Reparametrization &mMLP\\
Initial learning rate & 5e -3 \\
Decay rate &0.7 \\
Decay step &1000 \\
RBA decay parameter ($\gamma$)&0.999\\
RBA Learning rate ($\eta^*$)& 1e-2 \\
Offset ($\lambda_{o})$ & 0\\
ML framework & Jax \\
GPU &GeForce RTX $3090$ \\\hline
\end{tabular}
\label{p2:H1x4_ADF}
\end{table}

\begin{table}[H]
\centering
\caption{Training time for 2D Helmholtz ($a_1=1$, $a_2=4$)}
\begin{tabular}{
    >{\centering}m{5cm}
    >{\centering\arraybackslash}m{2.5cm}}
\hline
Method & Training time (ms/iteration) \\
\hline
Vanilla & 13.8\\
RBA & 13.8\\
RBA + mMLP& 18.6\\
RBA + Fourier& 22.6\\
mMLP + Fourier & 37.8\\
RBA + mMLP + Fourier & 34.4 \\
RBA + ADF& 14.2\\
mMLP + ADF & 18.4 \\
RBA + mMLP + ADF & 18.6 \\
\hline
\end{tabular}

\label{p2:t5_HM}
\end{table}
\subsection{3D Navier-Stokes settings}
\subsubsection{Collocation points and training groups generation}
\label{p2:AIV_implementation}
\begin{figure}[H]
 \centering
 \includegraphics[width=0.7\textwidth]{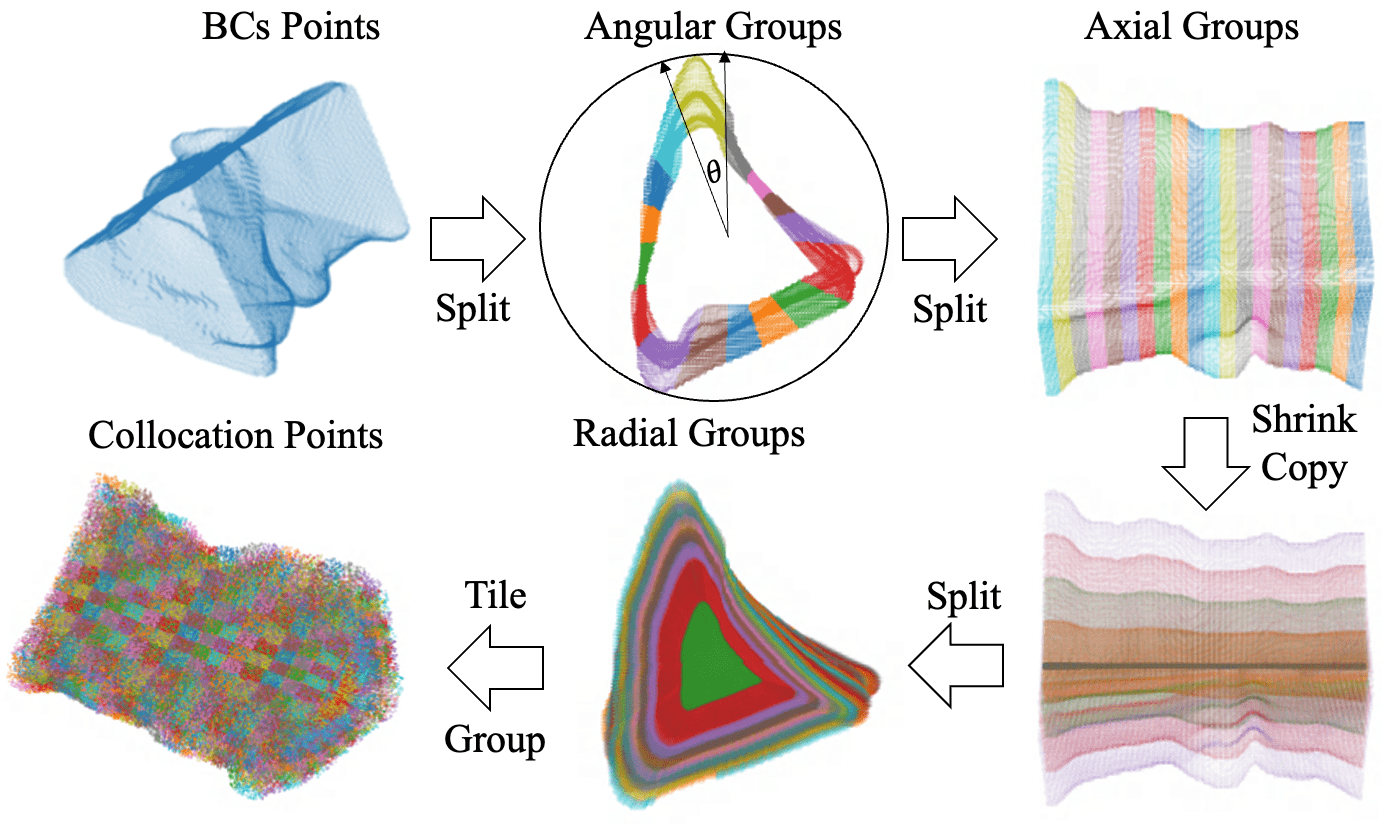}
 \caption[Collocation points generation for cerebral case]{\textbf{Collocation points generation:} We split our boundary data points into $N_{\theta}$ angular and $N_{y}$ axial subgroups. After that, we shrink-copy, combine the boundary points, and classify them in $N_{r}$ radial sections. Finally, we tile them and classify them into $N_t$ temporal subgroups and group the points into training groups based on their spatial and temporal coordinates.}
 \label{p2:fig:Datagen}
\end{figure}
We obtain our collocation points by shrinking and combining the boundary conditions as shown in Figure \ref{p2:fig:Datagen}. First, we split our boundaries into 25 angular $(N_{\theta})$ and 15 axial $(N_{y})$ subregions. Then, we shrink and combine them to get an equally spaced domain of 200 layers. After that, we split our domain into seven radial groups $(N_{r})$, tile them 101 times, and split the time domain into 20 temporal subregions $(N_{t})$. So, the total number of training  groups ($G_{c}$) for the NS would be:

$$G_{c}=N_t\times N_r \times N_{\theta} \times N_y=52500$$

Similarly, we split the boundary conditions with $N_{\theta}=25$, $N_{y}=15$ and $N_t=15$ as:

$$G_{b}=N_t\times N_{\theta} \times N_y=5625$$

Notice that the total number of collocation $G_{c}\times Nb =4.98e6$ and BC points $G_{b}\times Nb =5.34e5$ approximately matches the number of training points used in the previous study (i.e., $5e6$ and $5e5$ ) \cite{boster2023artificial}.
\subsubsection{Navier-Stokes approximation}

We approximate the 3D velocities ($u,v,w$) and pressure ($p$) as follows:
\begin{align*}
    u&=u_su_{NN}(t,x,y,z)\\
    v&=v_sv_{NN}(t,x,y,z)\\
    w&=w_sw_{NN}(t,x,y,z)\\
    p&=p_sp_{NN}(t,x,y,z)\text{,}
\end{align*}

\noindent where $t,x,y,z$ are the inputs (i.e., time and position), $u_{NN},v_{NN},w_{NN},p_{NN}$ are the neural network outputs and $(u_{s},v_{s},w_{s},p_{s})=(0.1,1,0.1,100)$. 
\begin{table}[H]
\centering
\caption{Global weights $w$ for AIV}
\begin{tabular}{cc}
\hline
Los term & Global weight \\
\hline
No-slip BC in $x$ &100\\
Non-slip BC in $y$ &100\\
Non-slip BC in $z$ &100\\
PTV velocity in $x$ &50\\
PTV velocity in $y$ &5\\
Conservation of Momentum in $x$ & 1\\
Conservation of Momentum in $y$  & 1\\
Conservation of Momentum in $z$ & 1\\
Conservation of Mass  & 10\\
\hline
\end{tabular}
\label{p2:t_Weights_AIV}
\end{table}

\begin{figure}[H]
 \centering
 \includegraphics[width=\textwidth]{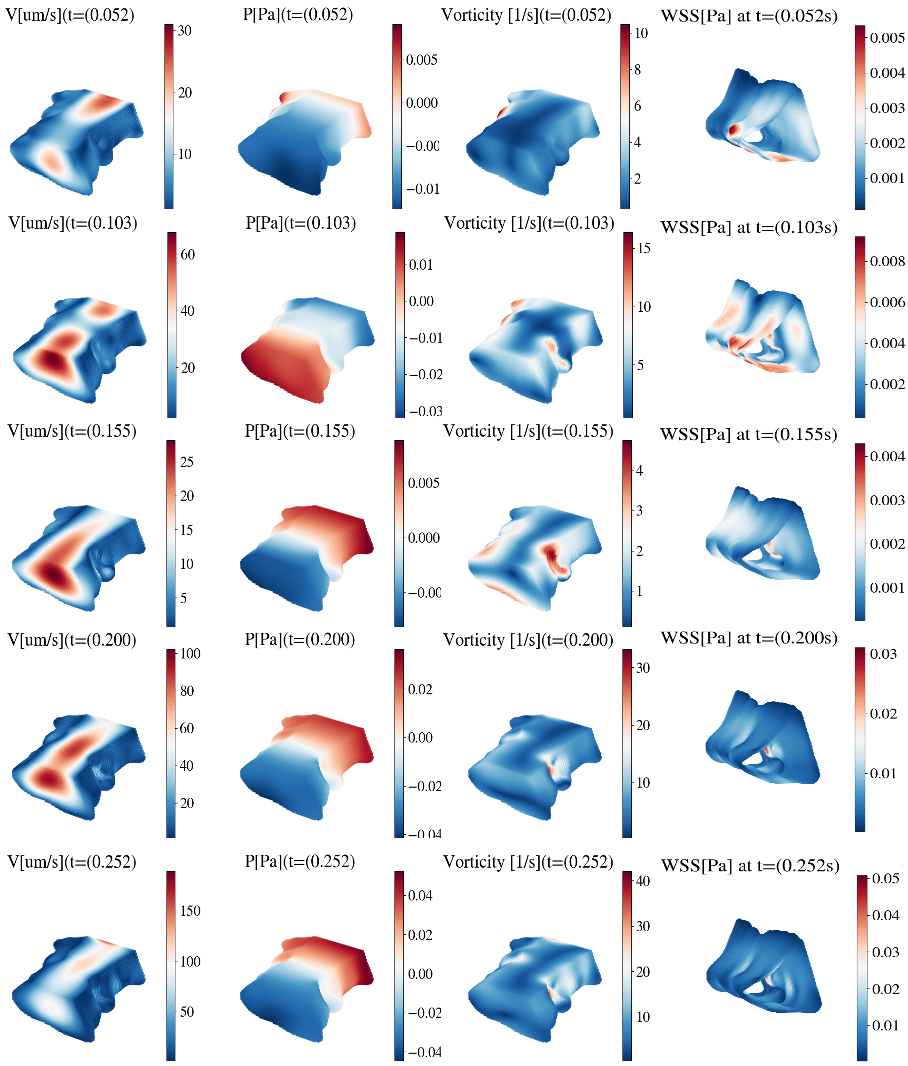}
 \caption[AIV Results]{\textbf{AIV Results.} Once the model is trained, we get a continuous and differentiable function for the velocity ($\mathbf{v}=[u,v,w]$) and pressure ($p$). Thus, following \cite{boster2023artificial}, we use automatic differentiation to compute the vorticity ($\boldsymbol{\omega}$) and wall shear stress (WSS). We present our results at five different time steps that describe the cardiac cycle $T$.}
 \label{p2:fig:Snap_AIV}
\end{figure}

Empirically, we observed that there is a performance improvement by assigning specific global weights to each loss term. The case-specific multipliers used for this problem are shown in Table~\ref{p2:t_Weights_AIV}. The remaining details about the implementation are described in Table~\ref{p2:AIV_imp}. Additional AIV results for different time steps along the cardiac cycle $T$ are shown in Figure~\ref{p2:fig:Snap_AIV}.
\subsubsection{Other implementation details}

\begin{table}[H]
\centering
\caption{AIV Implementation Details}
\begin{tabular}{ccc}
\hline
 \multicolumn{2}{c}{Implementation Details} \\\hline
Number of BCs data points  & 5.34e5 \\
Number of PTV data points (Train) & 9.4e3 \\
Number of PTV data points (Validation) & 1.4e4 \\
Number of collocation points & 4.98e6 \\
BCs batch size & 5.63e3\\
PTV data batch size & 9.40e3 \\
Navier-stokes batch size & 5.25e3 \\
Number of hidden layers & 8\\
Neurons per layer & 200\\
Activation function& sin\\
Reparametrization &WN\\
Initial learning rate & 2e-3 \\
Final learning rate & 1.5e-4 \\
Decay rate & 0.9\\
RBA decay parameter ($\gamma$)&0.999\\
RBA Learning rate for PTV and NS ($\eta^*$)& 2.0e-1 \\
RBA Learning rate for BCs ($\eta^*$)& 2.0e-2 \\
Offset ($\lambda_{o})$ & 1.0\\
ML framework & Jax \\
GPU &A100 \\\hline

\vspace{2em}

\end{tabular}
\label{p2:AIV_imp}
\end{table}
\vspace{-20pt}
\begin{table}[H]
\centering
\caption{Computational time for cases of the AIV ablation study.}
\begin{tabular}{
    >{\centering}m{4cm}
    >{\centering\arraybackslash}m{2.5cm}}
\hline
Method & Training time (ms/iteration) \\
\hline
Vanilla PINN& 325 \\
WN & 364 \\
RBA & 329 \\
RBA + WN& 366\\
\hline
\end{tabular}
\label{p2:tAIVspeed}
\end{table}

\newpage

 \bibliographystyle{elsarticle-num} 
 \bibliography{cas-refs}

\end{document}